%% file: main.tex
\documentclass[10pt,twocolumn,letterpaper]{article}

\usepackage[pagenumbers]{cvpr} 

\input{preamble}

\definecolor{cvprblue}{rgb}{0.21,0.49,0.74}
\usepackage[pagebackref,breaklinks,colorlinks,allcolors=cvprblue]{hyperref}

\def\paperID{} 
\def\confName{CVPR}
\def\confYear{2026}

\title{Artiverse: A Diverse and Physically Grounded Dataset for Articulated Objects}

\author{
Denys Iliash$^1$ \quad Jiayi Liu$^1$ \quad Egor Fokin$^1$ \quad Qirui Wu$^1$ \\
Ali Mahdavi-Amiri$^1$ \quad Manolis Savva$^1$ \quad Angel X. Chang$^{1,2}$\\
$^1$Simon Fraser University \quad $^2$Canada-CIFAR AI Chair, Amii \\
\href{https://3dlg-hcvc.github.io/artiverse/}
{\tt\small 3dlg-hcvc.github.io/artiverse/}
}

\begin{document}
\twocolumn[{%
\maketitle
\input{figures/teaser}
}]
\input{sec/0_abstract}
\input{sec/1_intro}
\input{sec/2_related_work}
\input{sec/3_annotation.tex}

\input{sec/4_dataset.tex}
\input{sec/5_experiments}
\input{sec/6_conclusion}

\subsubsection*{Acknowledgments}
This work was funded in part by a CIFAR AI Chair, a
Canada Research Chair, and  NSERC Discovery grants, and enabled by
support from the Digital Research Alliance of Canada.
We thank Weikun Peng, Han-Hung Lee, and Tristan Engst for helpful discussions, and Yiming Zhang for web interface support.
We thank our annotators for their dedication in ensuring the data quality: Arya Faghihy, Austin Wang, Daniel Chang, Derek Pun, Dingdong Yang, Dongchen Yang, George Katsadze, Han-Hung Lee, Ivan Tam, Jenna Lee, Jill Ference, Jiyeon Han, Junghun Byun, Max Vernikovskiy, Mingrui Zhao, Nina Dang, Qirui Wu, Saba Azzarouk, Simon Jin, Sonia Raychaudhuri, Stella Lin, Tommaso Galliena, Weikun Peng, Xiaohao Sun, Xingguang Yan, Yuefan Wu.

{
  \small
    \bibliographystyle{ieeenat_fullname}
    \bibliography{main}
}
\input{sec/7_suppl}


\end{document}

%% file: preamble.tex
\usepackage{times}
\usepackage{epsfig}
\usepackage{graphicx}
\usepackage{amsmath}
\usepackage{amssymb}
\usepackage{booktabs}
\usepackage{multirow}
\usepackage{comment}
\usepackage{xcolor}
\usepackage{comment}
\usepackage{array}
\usepackage{tikz}
\usepackage{listings}
\lstdefinelanguage{json}{
    morestring=[b]",
    morecomment=[l]{//}
}
\newif\ifshowcomments
\showcommentstrue 

\ifshowcomments

\else

\fi

\newcommand{\mypara}[1]{\noindent\textbf{#1}}
\newcommand{\denselist}{\itemsep 0pt\parsep=0pt\partopsep 0pt}

\usepackage{pifont}
\newcommand{\cmark}{\textcolor{ForestGreen}{\ding{51}}}%
\newcommand{\xmark}{\textcolor{pink}{\ding{55}}}%

%% file: figures/teaser.tex
\captionsetup{type=figure}
\includegraphics[width=\textwidth]{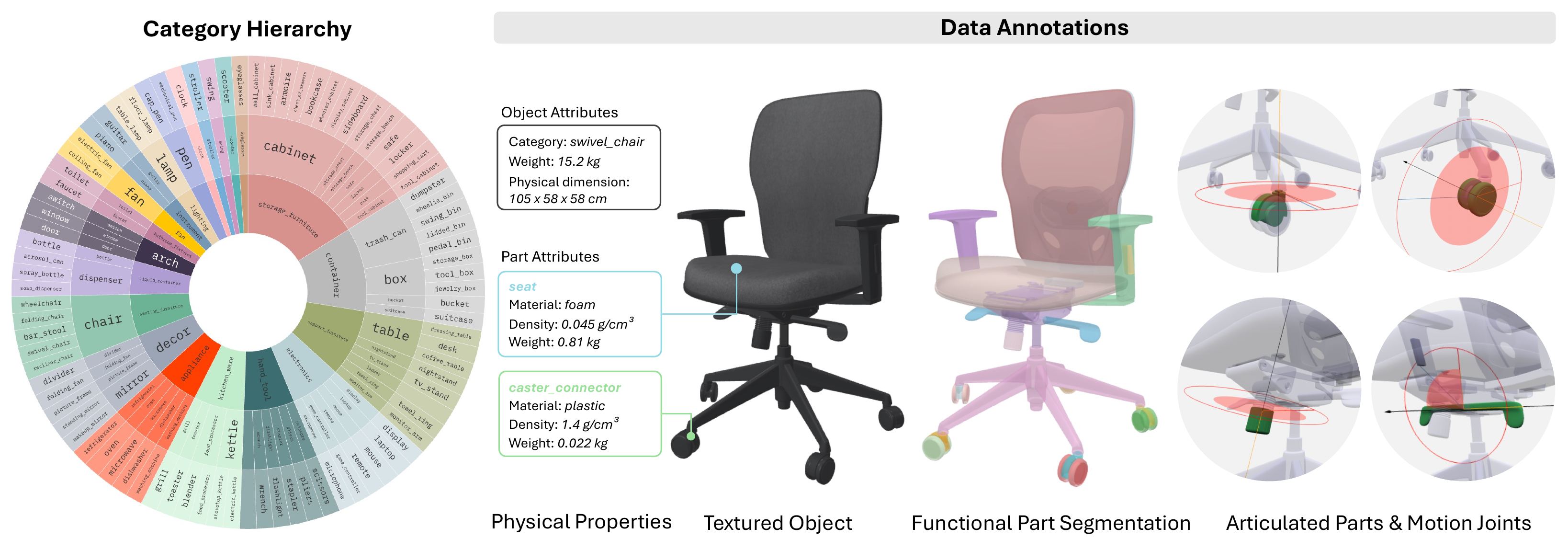}
\captionof{figure}{
We introduce Artiverse, a physically grounded 3D dataset of articulated objects across diverse categories, richly annotated with functional parts, motion joints for articulation, and physical properties including metric scale, per-part material, and mass.   
}
\label{fig:teaser}
\vspace{2em}

%% file: sec/0_abstract.tex
\begin{abstract}
We present Artiverse, a diverse and physically grounded dataset of high-quality articulated 3D objects designed for realistic functional modeling and simulation.
Artiverse contains 5.4K human-authored objects across a broad range of 88 categories, aggregated from multiple 3D static repositories.
Objects are annotated with functional parts, interior structures, realistic kinematic relationships and articulated joints including multi-DoF joints, and physical attributes such as metric scale, material, and mass.
We develop a semi-automated annotation pipeline that combines few-shot segmentation, geometric reasoning, and multi-stage human verification to achieve high-quality and efficient annotation, reducing manual annotation time by over 30\%.
We demonstrate the value of Artiverse on tasks of part mobility analysis, articulated object generation, and physics-based interaction.
Artiverse provides a data resource to advance functional understanding for articulated objects.
\end{abstract}

%% file: sec/1_intro.tex
\section{Introduction}
\label{sec:intro}

Objects are not merely static shapes. 
They are functional entities that serve a purpose in the world. 
A tabletop provides support, a cup holds liquids, and a hook enables hanging items. 
Beyond these general purposes implied by geometry, many everyday objects express their functionality through interaction: a drawer slides open to provide storage, a door swings to allow passage, and a faucet handle rotates to release water.
Understanding these objects requires modeling their \textit{functional interactivity}, which captures how objects are organized into movable parts, regions for interaction, and exhibit physically plausible behavior.

The notion of functional interactivity can be characterized along three complementary dimensions.
\textit{Functional parts} capture the structural decomposition of objects into components that contribute to their function, encompassing both movable and interactable elements.
\textit{Kinematic relationships} describe how articulated parts connect and move under joint constraints, and functional dependencies that enable coupled behaviors (e.g., open the microwave door by pushing a button).
\textit{Physical grounding and realism}, such as metric scale, mass and material, as well as detailed geometry and textures, ensure that articulated motions behave plausibly in simulation, faithfully reflecting the behavior of objects in the physical world.

Despite substantial progress in 3D dataset construction, existing resources fall short in capturing this level of functional realism. 
Large-scale object repositories~\cite{chang2015shapenet,deitke2023objaverse,deitke2024objaversexl} provide diverse geometry but mostly contain static shapes. 
Articulated object datasets~\cite{xiang2020sapien,liu2022akb} annotate part mobility but are limited in joint complexity (e.g., mostly single-DoF motions), diversity in part structure and category, and visual realism (e.g., simplified texture and missing interior geometries).
Physical properties are rarely modeled until recently, with ArtVIP~\cite{jin2025artvip} releasing physical digital replicas for 206 articulated objects, and PhysX-3D~\cite{cao2025physx} annotating physical attributes on PartNet mostly for rigid objects.
Thus, there are limited assets with sufficient geometric quality, diversity, and physical realism for studying functionally complete articulated objects at scale.

To bridge this gap, we introduce \textbf{Artiverse}, a dataset of 5,402 human-authored articulated objects across 88 diverse categories, richly annotated with functional parts, articulations, and physical attributes.
Objects are carefully filtered from multiple repositories (see \Cref{tab:ds_compare}).
Artiverse advances articulation modeling by supporting complex motion joints (e.g, cylindrical, universal joints) and modeling both kinematic and functional dependencies among parts and joints, enabling the study of chained and interdependent motions.
To ensure functional realism, we complete interior structures, scale objects to metric units, and estimate per-part mass.
All assets are standardized in orientation and released in consistent formats (e.g., GLB and USD), making them readily usable in simulation environments.

Building such a dataset at scale requires addressing the high annotation cost and the expertise needed for 3D geometry and kinematics.
We therefore design a semi-automated annotation pipeline integrating few-shot segmentation models and geometric reasoning to initialize proposals for segmentation, motion parameters, and physical attributes.
 Automated proposals are validated and corrected by human annotators in multiple stages throughout the pipeline, ensuring consistent and high-quality annotations, as illustrated in \Cref{fig:pipeline-overview}.
In practice, our pipeline reduces manual annotation time by 32.0\% and 33.5\% for segmentation and motion annotation respectively, with 50.12\% of the automatic annotations free of human correction, demonstrating efficiency and scalability.

Through its combination of kinematic complexity, geometric diversity, and physical grounding, Artiverse enables a broad range of research directions.
It offers reusable assets to build interactive scenes and immersive virtual environments for everyday activity understanding or interactive design.
It provides a rich resource for learning-based methods to advance functional understanding of articulated objects.
It supports generative modeling of articulated and physically grounded objects by providing denser coverage of the data distribution.
It can be used to train and evaluate embodied agents in physics-based simulators to facilitate policy learning for interactive tasks such as object manipulation.
In this paper, we illustrate some of these possibilities through benchmarks for part mobility analysis, experiments on generative modeling of articulated objects, and demonstrations of realistic interactions in simulation.
In summary, we make the following contributions:
\begin{itemize} \denselist
    \item We release Artiverse, a diverse 3D dataset of human-authored articulated objects, sourced from multiple high-quality shape repositories, and richly annotated with functional parts, articulation, and physical properties.
    \item We design a semi-automated annotation pipeline combining learning-based and geometric algorithms with human verification to efficiently produce high-quality annotations at scale.
    \item We show the value of Artiverse with benchmarks for part mobility analysis and generation of articulated objects, and demonstrations of realistic interactions in simulation.
\end{itemize}

\input{tables/dataset_compare}

%% file: tables/dataset_compare.tex
\begin{table*}[t]
    \centering
    \resizebox{\linewidth}{!}{
    \begin{tabular}{@{}llm{28em}ccccccrrr@{}}
    \toprule
                                           Dataset & Type     & Data Source                                          & Tex      & Kin & Fun & Mat    & Met & Mas & \# Obj & \# Part & \# Cat \\ \midrule
    
    RBO~\cite{martin2019rbo}                & Scan     & Self-scanned                                      & \cmark        & \xmark     & \xmark   & \xmark    & \cmark       & \xmark & 14         & 21            & 14           \\   
    ReArt-48~\cite{liu2022toward}           & Scan     & Self-scanned                                      & \cmark        & \xmark     & \xmark   & \xmark    & \cmark       & \xmark & 48         & -             & 5           \\
    AKB-48~\cite{liu2022akb}                & Scan     & Self-scanned                                      & \cmark        & \xmark     & \xmark   & \xmark    & \cmark       & \cmark & 2,037      & -             & 48           \\ \midrule
    RPMNet~\cite{yan2019rpmnet}             & Synth   & ShapeNet~\cite{chang2015shapenet}, 3D Warehouse   & \xmark        & \xmark     & \xmark   & \xmark   & \xmark       & \xmark & 969        & 1,420         & 43           \\
    Shape2Motion~\cite{wang2019shape2motion}& Synth   & ShapeNet~\cite{chang2015shapenet}, 3D Warehouse   & \xmark        & \xmark     & \xmark   & \xmark    & \xmark       & \xmark & 2,440      & 6,762         & 45           \\
    PartNet-Mobility~\cite{xiang2020sapien} & Synth   & PartNet~\cite{mo2019partnet}, 3D Warehouse        & \cmark        & \cmark     & \xmark   & \xmark    & \xmark       & \xmark & 2,346      & 11,753        & 46           \\
    GAPartNet~\cite{geng2023gapartnet}      & Both & PartNet-Mobility~\cite{wang2019shape2motion}, AKB-48~\cite{liu2022akb} & \cmark & \xmark & \xmark & \xmark & \xmark   & \xmark & 1,166      & 8,489         & 27           \\                               
    ACD~\cite{iliash2024s2o}                & Synth   & ABO~\cite{collins2022abo}, 3D-Future~\cite{fu20213d}, HSSD~\cite{khanna2024habitat} & \cmark & \cmark & \xmark & \xmark & \xmark & \xmark & 354    & 1,350     & 21 \\ 
    ArtVIP ~\cite{jin2025artvip} & Synth & Artist-created digital twins & \cmark & \cmark & \cmark* & \xmark & \cmark & \cmark & 205 & 705 & 29 \\ \midrule
    \textbf{Artiverse (ours)}                      & Synth   & ShapeNet~\cite{chang2015shapenet}, ABO~\cite{collins2022abo}, 3D-Future~\cite{fu20213d}, HSSD~\cite{khanna2024habitat}, Objaverse~\cite{deitke2023objaverse,deitke2024objaversexl}, TexVerse~\cite{zhang2025texverse}, 3DComPaT200~\cite{ahmed20253dcompat200}, PartNeXt~\cite{wang2025partnext}, SSDB~\cite{fisher2012example}  & \cmark & \cmark & \cmark & \cmark & \cmark & \cmark & \textbf{5,402} & \textbf{24,607} & \textbf{88} \\       
    \bottomrule
    \end{tabular}
    }
    \caption{
    Comparison of existing datasets containing 3D articulated objects.
    Columns indicate: \textbf{Tex}tured objects, \textbf{Kin}ematic and \textbf{Fun}ctional dependencies, per-part \textbf{Mat}erials, \textbf{Met}ric scale, per-part \textbf{Mas}ses, and numbers of articulated \textbf{obj}ects, \textbf{part}s (before completion), and \textbf{cat}egories.
    ``-'' indicates information not reported in the original paper, and ``*'' means partially available.
    Artiverse offers a comprehensive combination of functional part annotations, kinematic modeling, and physical attributes across diverse data sources and categories at scale.
    }
    \label{tab:ds_compare}
\end{table*}

%% file: sec/2_related_work.tex
\section{Related work}
\label{sec:related_work}
\mypara{Simulatable articulated assets.}
Existing articulated object datasets vary widely in geometry type, data source, annotation granularity, and scale, as summarized in \Cref{tab:ds_compare}.
Real scan datasets~\cite{martin2019rbo,liu2022toward,liu2022akb} reflect the realism of real-world objects but suffer from reconstruction noise and limited scalability due to the per-part capture effort.
More datasets are built on synthetic objects~\cite{wang2019shape2motion,yan2019rpmnet,geng2023gapartnet}.
PartNet-Mobility~\cite{xiang2020sapien} remains the most widely used.
ACD~\cite{iliash2024s2o} introduces more detailed objects and annotations but is limited to openable objects.
These datasets primarily focus on part mobility labels and overlook realistic textures or physical attributes.
Recent efforts begin to incorporate physical properties but remain limited in scale and coverage of articulated objects, with ArtVIP~\cite{jin2025artvip} releasing 206 articulated objects, and PhysX-3D~\cite{cao2025physx} mostly focus on rigid objects.
Building on these efforts, Artiverse contributes a comprehensive collection of articulated assets with broader category diversity, richer part structures, more complex motion patterns, and per-part materials and masses.

\mypara{Data annotation pipelines.}
Annotating articulated assets is highly labor-intensive, requiring expertise to ensure accurate and consistent motion specification.
Challenges include identifying movable part boundaries, defining plausible motion joints and ranges, and specifying coherent kinematic hierarchies.
Most existing efforts rely on fully manual annotation~\cite{wang2019shape2motion,xiang2020sapien,liu2022akb,geng2023gapartnet,jin2025artvip}, which limits scalability to diverse objects and categories.
Recently, PhysX-3D~\cite{cao2025physx} proposed a human-in-the-loop pipeline that uses vision-language models (VLMs) to infer physical attributes from 2D part renderings based on the fine-grained PartNet~\cite{mo2019partnet} hierarchy.
Without access to 3D geometry, purely 2D VLM-based annotation can be unreliable for tasks requiring spatial precision, such as segmenting parts on meshes and specifying joint axes for articulation.
To improve the accuracy and robustness of automatic proposals, we integrate multi-view few-shot models and geometry-based heuristics beyond pure VLMs to exploit the full 3D structure of each model.
This design allows our system to assist human annotators with minimal intervention while maintaining high-quality, consistent annotations at scale.

\input{figures/pipeline_overview}

\mypara{Articulated object modeling and applications.}
Articulated assets support a spectrum of research directions by providing realistic priors for learning, standardized benchmarks for evaluation, and interactive components for scene generation~\cite{yang2024physcene,xia2025drawer}, simulation~\cite{kerr2024robot,luo2025physpart}, and embodied AI tasks~\cite{ren2024infiniteworld,wang2025embodiedgen}.
Approaches for part mobility analysis focus on understanding the movable part structures and their articulation patterns from static 3D shapes~\cite{wang2019shape2motion,yan2019rpmnet,iliash2024s2o,goyal2025geopard} or 2D observations~\cite{jiang2022opd,sun2024opdmulti}, but achieving robustness across diverse objects remains challenging.
Reconstruction methods recover 3D articulated models from multi-view and/or multi-state images~\cite{liu2023paris, weng2024neural,liu2025building,guo2025articulatedgs,wu2025reartgs,deng2025paoli}, though handling occlusion and motion ambiguity remains an open problem.
Generative methods learn structural and motion prior from 3D~\cite{lei2023nap,gao2025meshart,kreber2025guiding} or metadata~\cite{liu2024cage,liu2024singapo,le2024articulate,wu2025dipo} to synthesize interactive parts, but are often constrained by limited diversity and realism of available training data.
Procedural methods synthesize articulated objects through parametric or functional recombination of parts~\cite{lian2025infinite,joshi2025infinigen,sun2025arti,guo2025kinematic}, though how to ensure the functional validity and consistency with real data distribution remains non-trivial.
These advances highlight the growing demand for large-scale, realistic, and physically-grounded articulated datasets.

%% file: figures/pipeline_overview.tex
\begin{figure*}[t]
\centering
\includegraphics[width=\linewidth]{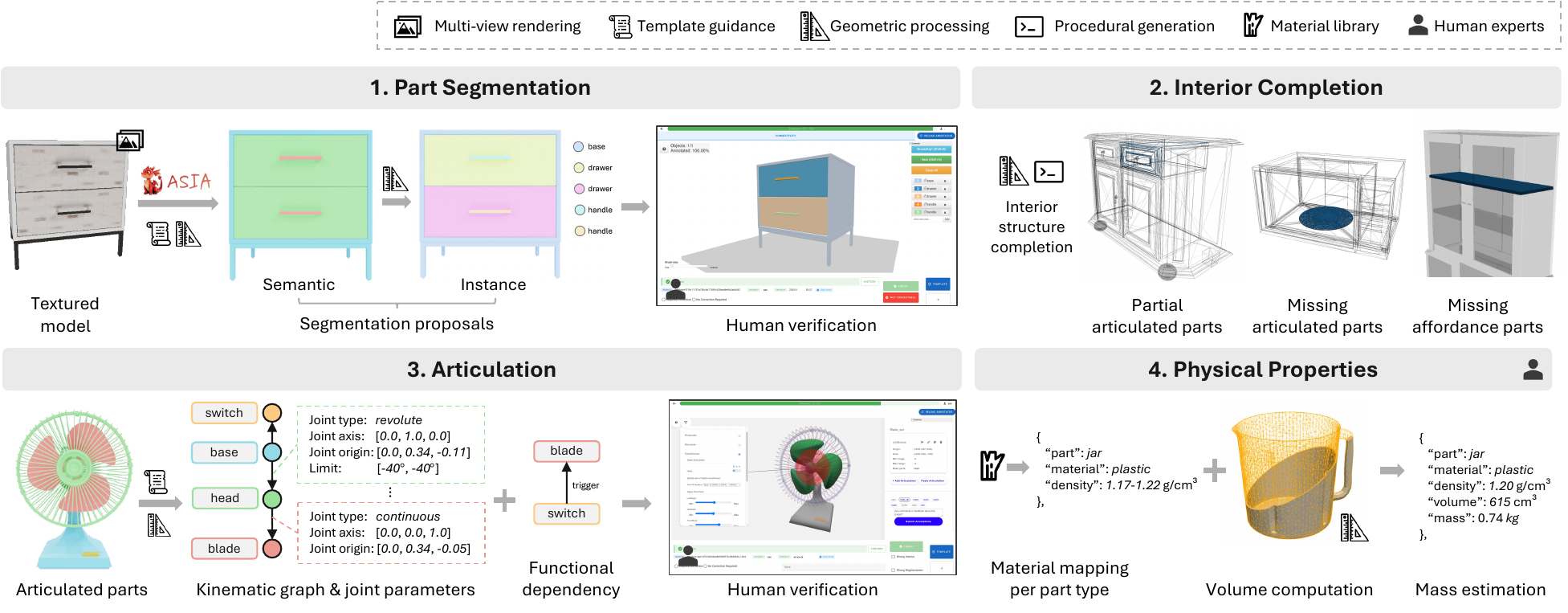}
\caption{
Overview of the key modules in our semi-automated annotation framework.
Each selected object progresses through four stages sequentially after preprocessing: 
1) functional part segmentation, 
2) optional interior completion for objects with missing internal structures, 
3) annotation of articulation mechanisms, and
4) estimation of physical properties.
Human verification is incorporated at the end of the segmentation and articulation stages to correct errors and ensure that accurate annotations are passed to subsequent modules.
After a final verification at the end of the pipeline, the output is a fully annotated 3D articulated object ready for simulation.
}
\vspace{-1em}
\label{fig:pipeline-overview}
\end{figure*}

%% file: sec/3_annotation.tex
\section{Data Annotation}

\subsection{Design Desiderata}
\label{subsec: design_principles}

The construction of Artiverse is guided by the following considerations shaping our data selection and annotation.

\mypara{Object coverage.}
Artiverse focuses on human-made articulated objects for daily activities, such as furniture, tools, and appliances.
We prioritize categories appearing frequently in human environments (e.g., household, office, and outdoor) to ensure relevance for real-world applications.
We organize the objects in three levels of semantic categories: super-category, main category, and sub-category.
Super-categories are broad and general (e.g., appliance, lighting, kitchenware).
Main categories are specific to the object function (e.g., toaster, window, dispenser).
Sub-categories capture fine-grained distinctions in the usage context (e.g, office chair vs. bar stool).

\mypara{Functional part specification.}
We define a unified taxonomy of functional part labels that capture both articulated (e.g., levers, lids) and interaction-related (e.g., handles, backrest, shelves, racks) components.
These labels are designed to be category-consistent and reflect the functional roles of parts in enabling articulation and interaction.

\mypara{Annotation consistency.}
To maintain coherent semantics across the dataset, we use category-level templates that specify part labels, motion types and dependencies for each object family.
The templates are defined by human experts and are flexible to accommodate extra information discovered by annotators.
These templates guide both automated inference and human verification, ensuring consistent labeling and complete articulation coverage.

\subsection{Semi-automated Workflow}
We design a semi-automated annotation workflow, which includes the steps of model selection and preprocessing, part segmentation, interior completion, motion annotation, physical property estimation, and multi-stage human verification in between.
The key steps of the pipeline are illustrated in \Cref{fig:pipeline-overview}.
In the following sections, we discuss how we design each key module.
Please refer to the supplement for more implementation details.


\subsubsection{Part Segmentation}
Each object in Artiverse is decomposed into parts that are associated to an articulation hierarchy and interaction affordance.
For every object category, human experts design a template listing possible part labels shared across instances in the category.
These templates may be refined during the human verification stage if additional parts are discovered.
Leveraging the template, our part annotation strategy follows a two-stage process: we first generate initial part proposals using learning-based models and geometric algorithms, and then refine them through human verification.

\mypara{Initial proposals.}
Our part labels are designed to be semantically meaningful.
However, unlike general semantic segmentation, part boundaries are determined by functional behavior rather than purely visual or semantic similarity, leading to subtle but critical distinctions.
Therefore, it is essential to employ a segmentation model that can incorporate contextual definitions of each label, making few-shot models~\cite{liu2023partslip,kim2024partstad,thai20243,perla2025asia} a better fit than non-adaptable 3D segmentors~\cite{abdelreheem2023satr,decatur20233d,zhong2024meshsegmenter,decatur20243d}.
We opt for the state-of-the-art model ASIA~\cite{perla2025asia} in our pipeline, as it supports label customization in a few-shot manner via visual exemplars.
We train ASIA on groups of categories that share similar part structures and labels to minimize training redundancy.
We manually annotate a dozen representative shapes in 3D and render part masks from multiple views to serve as training data.
Once trained, ASIA produces label-aware part masks on multi-view images, reflecting our functional context.

\mypara{Map proposals to 3D part instances.}
To project 2D segmentation results onto 3D meshes, we first precompute an oversegmentation based on the mesh's topological connectivity.
Each over-segment aggregates label votes from multi-view projections according to the number of covered pixels.
Unlabeled internal surfaces are assigned part labels using a distance-based propagation that considers the physical proximity of sampled surface points.
Once semantic segmentation converges, part instances are decomposed from each semantic group.
We use a union-find algorithm to obtain disjoint connected sets of mesh segments, each forming an independent part instance, determined by local geometric continuity and spatial distance.

\mypara{Human verification.}
At the end of the segmentation stage, human verification checks the accuracy of part instances and labels proposed by the automated process above.
We develop a web interface for annotators to check whether corrections are required, and correct any mistakes with a 3D part annotation UI, refining part labels if needed (see the example interface in \Cref{fig:pipeline-overview}).
As the last step, human experts verify the annotations from each annotator to ensure clean and consistent part segmentation.


\subsubsection{Interior Completion}
For some categories such as appliances and furniture, the interior serves as a critical structure enabling the intended functionality.
However, the vast majority of objects in static datasets lack sufficiently detailed interior geometry.
To provide assets that behave realistically in physical simulation, we complete the interiors of these objects where necessary.
We focus on three types of interiors to complete.

\mypara{Partial articulated parts.}
Some articulated parts are partially modeled, with interior geometry omitted because it is not visible when the part is in the resting state.
This issue is typical for components exhibiting translational motion, where hidden regions are revealed only during articulation.
For example, many drawers are modeled with only the front panel, while the inner box is missing.
We design a geometry-informed completion algorithm for such parts, extending the geometry consistently with the motion constraints to ensure plausibility in full motion range.

\mypara{Missing articulated parts.}
Certain appliances contain movable internal components that are entirely absent in many 3D assets.
Examples include dishwashers with baskets for dishes, microwaves with rotating turntables, and refrigerators with movable storage compartments.
To address this, we implement a procedural algorithm that populates these missing elements by referencing similar parts from other objects within the same category, ensuring functional completeness and structural realism.

\mypara{Missing interaction affordance parts.}
Furniture and appliances that provide storage space are often modeled as hollow shells, resulting in unrealistic behavior when interacting with other objects.
To restore functional realism, we add interior structures such as shelves, racks, and dividers that correspond to the intended use of the object, e.g., refrigerator shelves for food placement or wardrobe racks for hanging clothes.
These interior completions are generated procedurally following category-specific templates that ensure geometric consistency and plausible spatial layout.


\subsubsection{Motion Annotation}
At this stage, models have verified part segmentations and complete geometry, including interiors.
We follow a similar strategy as in part segmentation, where automated motion proposals are first generated and later verified by human experts to ensure accuracy and physical plausibility. 

\mypara{Initial proposal.}
We design algorithms to predict the joint parameters, including joint type, motion axis, and joint limits.
In particular, we establish general rules that use PCA-aligned oriented bounding boxes for each part and sampled points around contact regions to infer motion parameters for different motion types, constrained by geometric contact and collision relationships.
Kinematic dependencies between articulated parts are then inferred based on spatial connectivity and dependency options defined in the per-category templates.
The inference rules are reusable across categories and are registered for different part types, requiring only minor adjustments for specialized components.

\mypara{Human verification.}
In the motion verification stage, human annotators validate two key aspects: 1) whether the completed interior geometry is realistic; and 2) whether the proposed motions are physically plausible.
We develop a web-based interface that enables annotators to visualize and adjust articulations interactively, modify joint parameters, and edit kinematic graphs as needed (see \Cref{fig:pipeline-overview} for an example page).
To improve efficiency, the interface supports features such as motion copying between similar parts and instant motion previews, allowing rapid correction and validation of proposed motion parameters.

\input{figures/assets_qual}

\subsubsection{Physical Properties}
To support physics-based simulation, we annotate objects with metric scale, and parts with materials and masses.

\mypara{Metric scale}.
The metric scale is either obtained from the original data source or estimated during preprocessing.
We ask LLMs to infer a reasonable range of physical dimensions for each sub-category, and then sample within the range to rescale the object accordingly.
It is critical to estimate metric scale in early stages of the pipeline, as our geometric algorithms for part instance decomposition and interior completion rely on this for robustness.

\mypara{Per-part mass}.
Each part's mass is estimated as the product of its approximated volume and the density sampled within the range of an estimated material.
We predefine a list of material labels and query LLMs for a reasonable density range for each material.
We assign each part label a default material label as the initial proposal, and later verify it manually in the end. 
Accurate volume estimation is non-trivial for surface meshes, as our data is mixed with solid parts enclosed by shells and hollow parts lacking modeled thickness.
Thus, we design different processing strategies for hollow and solid parts, guided by their semantic labels.
For solid parts, we extract a volumetric mesh by tetrahderalization and compute the volume directly.
For hollow parts, we offset the surface inwards along the normal to create a shell, and then tetrahedralize to approximate the volume.

%% file: figures/assets_qual.tex
\begin{figure}[t]
\centering
\includegraphics[width=\linewidth]{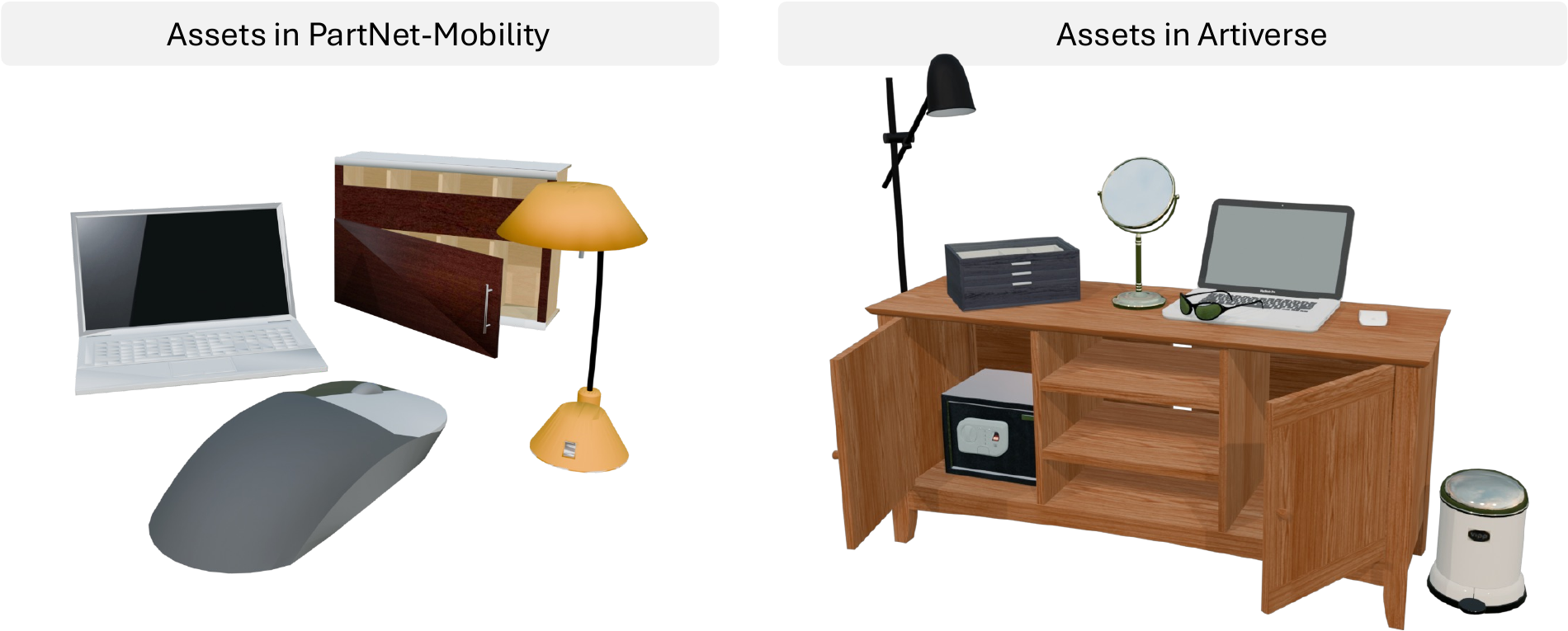}
\caption{
Visual comparison of assets from Artiverse and PartNet-Mobility~\cite{xiang2020sapien}.
We manually assemble sub-scenes using assets from both datasets with the original object scales preserved.
Artiverse objects exhibit higher geometric fidelity, realistic textures, and consistent metric scale, resulting in assets that are visually richer and more suitable for realistic simulation and scene design.
}
\label{fig:asset_qual}
\end{figure}

%% file: sec/4_dataset.tex
\section{Artiverse}
\subsection{Data Statistics}
\label{subsec:data_statistics}
Artiverse contains 5,402 human-made articulated objects sourced from 10 repositories, covering 20 super-categories, 61 main categories, and 88 sub-categories, encompassing a wide range of furniture, appliances, tools, and everyday household objects.
\Cref{fig:distributions} summarizes the distribution of data sources, object counts per super-category, and different joint types.
\Cref{tab:artiverse_stats} reports a detailed breakdown of our annotation statistics in comparison with two prior datasets: PartNet-Mobility~\cite{xiang2020sapien} (PM) and ArtVIP~\cite{jin2025artvip}.
Artiverse provides broader category coverage with more objects overall, more functional parts, and more varied and complex articulations per object.
Together, these statistics highlight the diversity and functional complexity captured in Artiverse, which can benefit downstream applications.
In \Cref{fig:asset_qual}, we qualitatively compare our assets with PM by manually assembling sub-scenes, showing that Artiverse provides data with higher geometric fidelity, realistic textures,
and consistent metric scale, resulting in assets that are visually
richer and more suitable for realistic simulation and scene design.

\input{tables/artiverse_stats}

\subsection{Data Schema}

Each object in Artiverse is represented as a textured 3D mesh enriched with a hierarchical annotation schema that captures its structural, functional, and physical properties.
Annotations are organized in two levels:
(1) the \textit{object level}, which stores global metadata such as unique identifier, category labels, dataset source, and metric scale;
(2) the \textit{part level}, which encodes articulated and interaction-related components, their semantic labels, motion parameters, material properties including label and density, and mass.

All assets are released in multiple interoperable formats.
The textured geometry is provided in glTF binary (GLB) for visualization and rendering, Universal Scene Description (USD) and URDF for simulation compatibility, and JSON metadata for easy parsing and programmatic analysis.
Each object has a standardized orientation, consistent scale, and unified coordinate convention, enabling direct integration into physics simulators.
\Cref{subsec:artiverse_in_simulation} showcases the usage of Artiverse assets in simulators.

\input{figures/distributions}

\subsection{Annotation Efficiency}

We evaluate the efficiency of our semi-automated annotation pipeline by comparing it against a fully manual annotation process for the two main annotation stages: part segmentation and motion annotation.
For each category, we sample 5 representative objects and record the time used by annotators to complete each stage manually, and compare it with the human-correction time logged through our annotation interface.
Using this protocol, our automatic proposals can save 32.0\% and 33.5\% of the manual annotation time for segmentation and motion annotation, respectively.
In our semi-automated pipeline, the average human correction time is reduced to 1.5 minutes for part segmentation and 1.3 minutes for motion annotation.
We also record that 50.12\% of the parts on average do not require human adjustment afterwards.
To ensure consistency across annotators, an expert verification pass is performed on the corrected results, which adds an additional 0.8 minutes per object on average.

These measurements show that our semi-automated pipeline achieves a substantial reduction in manual annotation effort, making it feasible to annotate thousands of objects while maintaining high-quality and consistent outputs.
The modular design of the pipeline also allows the integration of improved automated components in the future, enabling gains in scalability and annotation reliability.

%% file: tables/artiverse_stats.tex
\begin{table}[t]
    \centering
    \resizebox{\linewidth}{!}{
    \begin{tabular}{@{}lrrrrrrrrr @{}}
    \toprule
    &  & \multicolumn{2}{c}{Category}  & \multicolumn{2}{c}{\# Func. Parts} & \multicolumn{2}{c}{\# Arti. Parts} & \multicolumn{2}{c}{\# Joints} \\
    \cmidrule(lr){3-4} \cmidrule(lr){5-6} \cmidrule(lr){7-8} \cmidrule(lr){9-10}
    Dataset  & \# obj & Total & Avg \# obj & Total & Avg & Total & Avg & 1-DoF & 2-DoF \\
    \midrule

    PM & 2,346 & 46 & \textbf{51.0}  & 14,100 & 6.0 & 11,753 & 5.0 & 11,753 & 0 \\
    ArtVIP & 205 & 29 & 7.1 & 1,784  & 8.8 & 705 & 3.4 & 705 & 0 \\
    Artiverse & \textbf{5,402} & \textbf{88} & 49.4  & \textbf{38,608} & \textbf{8.9} & \textbf{24,607} & \textbf{5.4} & \textbf{23,640} & \textbf{480} \\

    \bottomrule
    \end{tabular}
    }
    \caption{Comparison of articulation annotation statistics.
    Artiverse exhibits broader category coverage, richer functional part structure, and more complex articulation on average. 
    }
    \label{tab:artiverse_stats}
\end{table}

%% file: figures/distributions.tex
\begin{figure}[t]
\centering
\includegraphics[width=\linewidth]{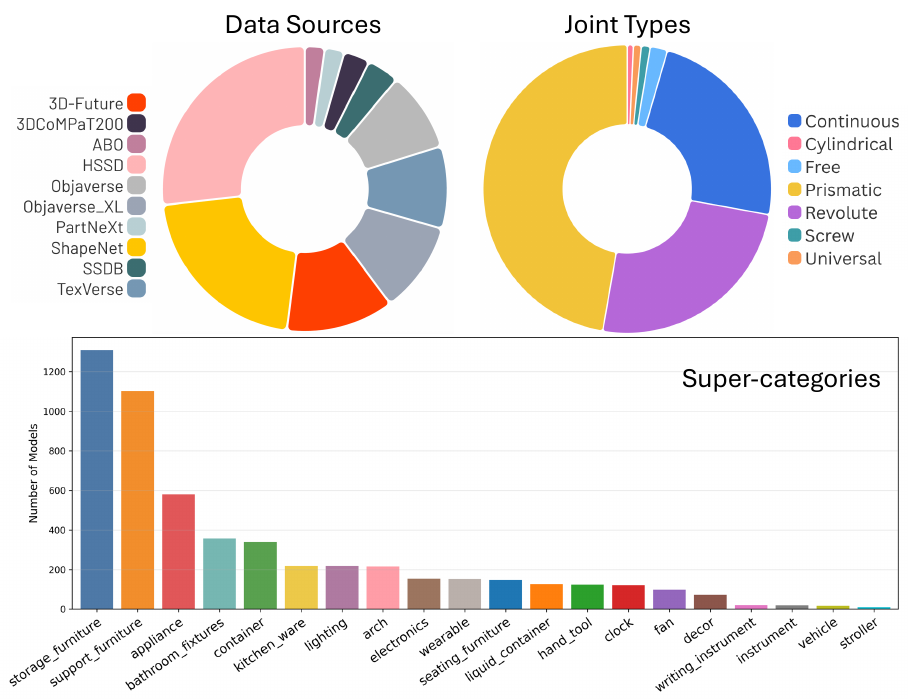}
\caption{
We visualize the distribution of objects across super-categories and across data sources, and the frequency of different joint types.
Artiverse spans a broad range of object families with balanced coverage and aggregates assets from multiple repositories to increase model and category diversity.
The distribution of joint types highlights the presence of common 1-DoF and complex multi-DoF joints (e.g., universal), reflecting motion diversity.
} 
\label{fig:distributions}
\end{figure}

%% file: sec/5_experiments.tex
\section{Experiments}
To demonstrate the utility and versatility of Artiverse, we evaluate its effectiveness across multiple downstream applications that rely on 3D articulated objects.
We focus on two representative benchmark tasks: \textit{part mobility analysis} and \textit{articulated object generation}.
These tasks capture complementary aspects of reasoning and synthesis that both benefit from dataset scale and diversity.
Part mobility analysis (\Cref{subsec:exp_estimation}) assesses how well a method can infer motion types, joint parameters, and articulated structures from geometric cues, reflecting the informativeness and precision of our articulation annotations.
Articulated object generation (\Cref{subsec:exp_generation}) evaluates the ability to synthesize functionally consistent articulated shapes, highlighting the dataset's diversity and annotation completeness.
Finally, we demonstrate the simulation readiness of Artiverse assets through physics-based simulation (\Cref{subsec:artiverse_in_simulation}), illustrating their usability for embodied AI and interactive environments.

\input{tables/quant_segment_mot_s2o}

\subsection{Part Mobility Analysis}
\label{subsec:exp_estimation}

\mypara{Task definition.}
This task measures both geometric understanding and the ability to recover motion patterns from static geometry.
Given a static 3D shape, the goal is to decompose the object into articulated part segmentation and predict the joint parameters associated with each part.

\mypara{Setup.}
We use a recent method FPNGroupMot proposed in S2O~\cite{iliash2024s2o} 
to benchmark the task performance trained on Artiverse against the prior dataset PartNet-Mobility~\cite{xiang2020sapien} (PM).
The models take as input a point cloud sampled from the object surface and output instance segmentation of articulated parts and their corresponding motion parameters, including motion type, joint axis, and origin.
We split both datasets into train/val/test sets with an 80/10/10 ratio.
The model is trained on each dataset respectively, and evaluated on both test sets to show the generalization capability.

\mypara{Metrics.}
We break down the task evaluation into articulated part segmentation and motion prediction.
We follow the evaluation metrics used in S2O.
Specifically, we use precision (P), recall (R), and F1 score as segmentation metrics with area-based instance matching macro-averaged over the object.
For motion estimation, we report macro F1 scores with matches further refined by motion type (+M), motion type and axis (+MA) and origin (+MAO).

\mypara{Discussion.}
The quantitative results are in \Cref{tab:quant_segment_mot_s2o}.
When benchmarking with PM, both methods achieve significantly better performance when trained on Artiverse than when trained on PM, demonstrating that the diversity and comprehensive annotations of our dataset enhance generalization of existing models.
However, when evaluated on the Artiverse test set, performance drops noticeably compared to prior benchmarks, indicating that the higher geometric fidelity, richer articulation patterns, and complex multi-DoF or dependent joints in Artiverse present new challenges.
This suggests that while current approaches can exploit larger data for improved learning, they still struggle to reason over more complex and variable geometric and motion structures.
We believe Artiverse therefore not only serves as a stronger training resource but also establishes a more demanding benchmark that exposes the limitations of existing methods and motivates the development of models capable of unified reasoning over articulation behaviors.
\input{tables/quant_singapo_acd}

\subsection{Articulated Object Generation}
\label{subsec:exp_generation}

\mypara{Task definition.}
The goal of articulated object generation is to synthesize new 3D shapes that are both geometrically realistic and functionally coherent, producing plausible articulated parts, motion joints, and their kinematic dependencies.
We benchmark this task under a conditional generation setting, where the model receives a single-view image of an object as input and generates corresponding articulated shapes that are structurally and geometrically consistent with the input.
This setup provides a strong test of generalization, as it requires the model to infer hidden geometry, articulation, and motion behavior in a constrained setting from limited visual information, particularly for in-the-wild or previously unseen inputs.

\mypara{Setup.}
We benchmark two recent open-sourced methods on the image-conditioned generation task: SINGAPO~\cite{liu2024singapo} and Articulate-Anything~\cite{le2024articulate}.
SINGAPO learns high-level articulation structure from training data, while Articulate-Anything leverages VLMs and retrieval of objects and parts from a database.
We report SINGAPO performance trained on Artiverse against PM to show the effect of the more diverse data in Artiverse.
Articulate-Anything serves as a complementary baseline that shows how well a VLM prior can generalize articulation understanding without fine-tuning on annotated data.
The original SINGAPO method leverages a diffusion model to generate part layout, kinematic structure, and part labels to retrieve part meshes from the library to produce the final 3D geometry.
We retrain SINGAPO on the full Artiverse, excluding objects with more than 32 parts, and evaluate on PM and our test split.

\mypara{Metrics.}
We follow the evaluation setup from SINGAPO to evaluate the structural and geometric consistency of generated objects against ground truth data.
The metrics are defined using 3D bounding box IoU (gIoU) parts at resting state (RS-) or across multiple articulated states (AS-) averaged over objects.
Collision rate during articulation is measured using the averaged overlap ratio (AOR).

\input{figures/singapo_qual}

\mypara{Discussion.}
The qualitative results are shown in \Cref{fig:singapo_qual} and quantitative results are shown in \Cref{tab:quant_singapo_acd}.
We observe that the SINGAPO model trained on Artiverse generalizes well overall across the test set.
From the metrics of AS-gIoU and AOR, we see that Articulate-Anything still lags behind SINGAPO trained on Artiverse, indicating the limitation of VLM priors in capturing fine-grained articulation details without direct supervision.
Overall, these results suggest that Artiverse serves as a valuable resource for training generative models that can produce functionally plausible articulated objects, and that learning from structured data priors remains crucial for detailed articulation synthesis.

\input{figures/simulation}

\subsection{Artiverse in Simulation}
\label{subsec:artiverse_in_simulation}

Artiverse provides simulation-ready articulated assets in both URDF and USD formats, enabling seamless integration with a wide range of physics engines and robotics simulators. To demonstrate the practical simulation readiness, physical realism, and articulation correctness of our assets, we load them directly into the Genesis~\cite{Genesis} and train a policy to interact with the example assets.
\Cref{fig:simulation} illustrates an interaction with a cabinet and successfully opens one of its drawers, showcasing accurate joint modeling.

%% file: tables/quant_segment_mot_s2o.tex
\begin{table}[t]
\centering
\resizebox{\linewidth}{!}{
\begin{tabular}{@{} ll rrr rrr @{}}
\toprule
& & \multicolumn{3}{c}{Segmentation} & \multicolumn{3}{c}{Motion} \\
\cmidrule(lr){3-5} \cmidrule(lr){6-8}
Eval  & Train  & P$\uparrow$ & R$\uparrow$ & F1$\uparrow$ & +M$\uparrow$ & +MA$\uparrow$ & +MAO$\uparrow$ \\
\midrule
\multirow{2}{*}{PM} & PM & 81.8 & 46.0 & 54.2 & 22.1 & \textbf{17.0} & \textbf{14.3}  \\
                    & Artiverse & \textbf{82.2} & \textbf{47.9} & \textbf{55.8} & \textbf{22.4} & 15.9 & 8.7  \\ \midrule
\multirow{2}{*}{Artiverse} & PM & 72.8 & 31.7 & 40.6 & 7.2 & 2.5 & 1.2 \\
                           & Artiverse & \textbf{77.0} & \textbf{43.0} & \textbf{50.7} & \textbf{22.4} & \textbf{16.6} & \textbf{10.8} \\ 
\bottomrule
\end{tabular}  
}
\caption{
We benchmark the task of part mobility analysis with a recent method FPNGroupMot~\cite{iliash2024s2o} in a cross-dataset evaluation between Artiverse and PartNet-Mobility (PM), training on each dataset respectively.
The results show that training on Artiverse consistently outperforms training on PM in both segmentation and motion prediction metrics.
This demonstrates that the superior diversity and quality of Artiverse enhances model generalization.
}
\label{tab:quant_segment_mot_s2o}
\end{table}

%% file: tables/quant_singapo_acd.tex
\begin{table}[t]
  \centering
  \label{tab:quant_singapo_acd}
  \resizebox{\linewidth}{!}{%
    \begin{tabular}{@{}llrrrrr@{}}
      \toprule
      & & \multicolumn{4}{c}{Reconstruction quality $\downarrow$} & \multicolumn{1}{c}{Collision} \\
      \cmidrule(lr){3-6} \cmidrule(lr){7-7}
      Eval & Methods & RS-$d_{\text{gIoU}}$ & AS-$d_{\text{gIoU}}$ & RS-$d_{\text{cDist}}$ & AS-$d_{\text{cDist}}\downarrow$ & AOR$\downarrow$ \\
      \midrule
      \multirow{2}{*}{PM} 
          & SG ~\cite{liu2024singapo} & \textbf{0.756}& \textbf{0.768}& \textbf{0.178}& \textbf{0.229}& \textbf{0.022}\\
          & AA~\cite{le2024articulate} & 1.172& 1.179& 0.425& 0.780& 0.024\\ \midrule
      \multirow{2}{*}{Artiverse} 
          & SG ~\cite{liu2024singapo} & \textbf{0.810}& \textbf{0.822}& \textbf{0.105}& \textbf{0.158}& \textbf{0.009}\\
          & AA~\cite{le2024articulate} & 1.250& 1.258& 0.385& 0.856& 0.042\\ 
      \bottomrule
    \end{tabular}%
  }
  \caption{
    We benchmark the task of image-conditioned articulated object generation with Articulate-Anything~\cite{le2024articulate} that purely relies on VLMs, and an adapted version of SINGAPO~\cite{liu2024singapo} re-trained on Artiverse.
    Across both evaluation sets, SINGAPO achieves consistently higher performance, indicating the limitations of VLM priors for modeling 3D articulated structures and highlighting the benefit of Artiverse data in distribution modeling.
  }
\end{table}

%% file: figures/singapo_qual.tex
\begin{figure}[t]
\centering
\includegraphics[width=\linewidth]{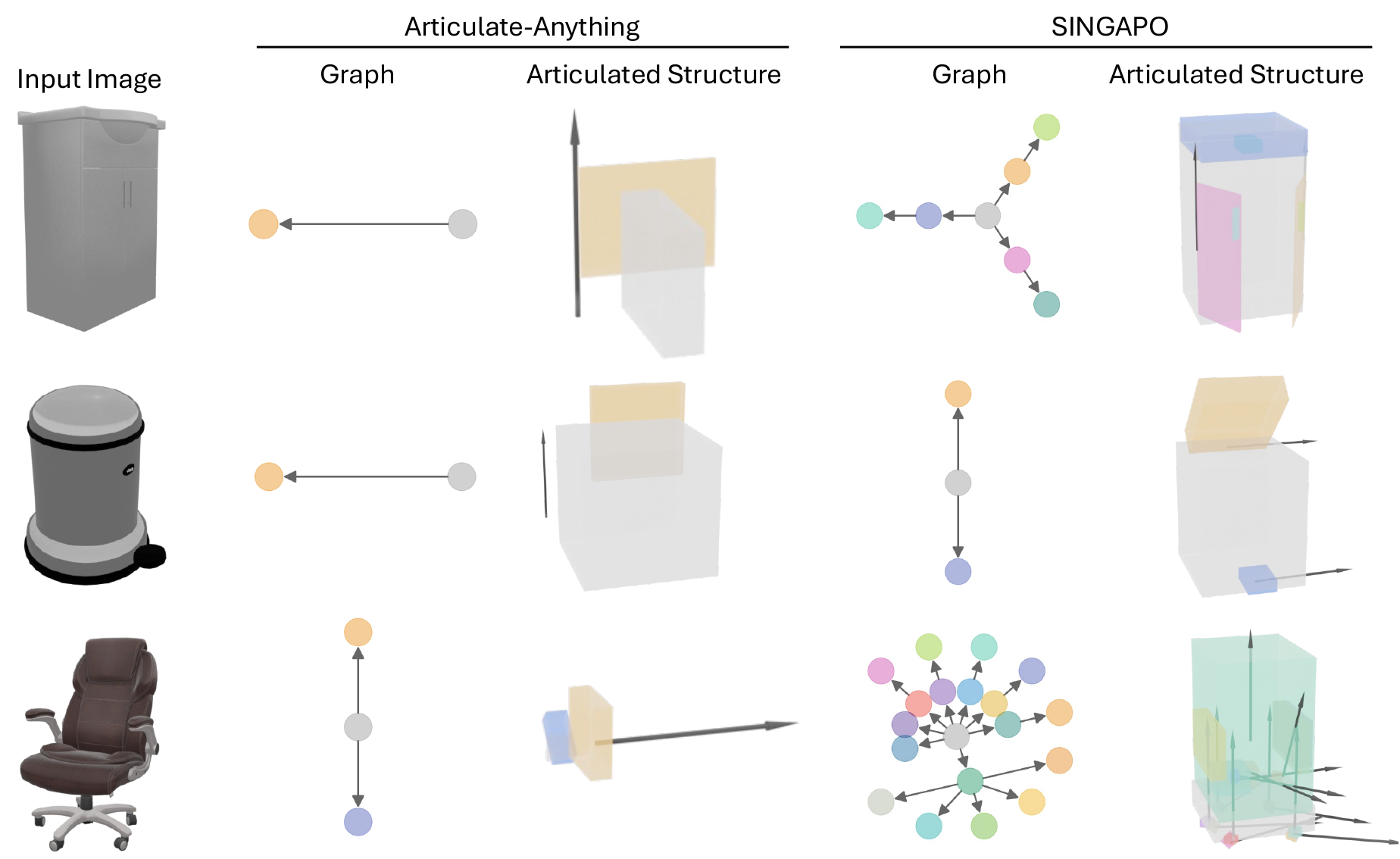}
\caption{Qualitative comparison on the task of image-conditioned generation of articulated objects by showing the generated kinematic graph and structures.
SINGAPO trained on Artiverse outperforms Articulate-Anything that relies on the VLM knowledge only across different categories.
}
\label{fig:singapo_qual}
\end{figure}

%% file: figures/simulation.tex
\begin{figure}[t]
\centering
\includegraphics[width=\linewidth]{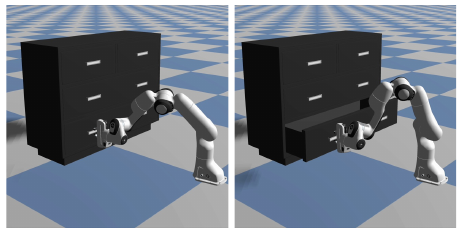}
\caption{Loading an example Artiverse asset into Genesis~\cite{Genesis} to show realistic and accurate joint modeling. 
}
\label{fig:simulation}
\end{figure}

%% file: sec/6_conclusion.tex
\section{Conclusion}
We presented Artiverse, a comprehensive dataset of physically grounded and articulated 3D objects. 
It collects human-made everyday objects across diverse categories, each annotated with detailed functional part structure, articulation mechanisms, and physical attributes such as material, mass, and metric scale.
To construct the dataset efficiently, we developed a semi-automated annotation pipeline that integrates geometric reasoning, few-shot segmentation, and human verification in multiple stages.
This pipeline provides a scalable and flexible framework for producing high-quality annotations across heterogeneous 3D assets.
Through benchmarks on part mobility analysis and articulated object generation, we show that Artiverse improves the robustness and generalization of existing approaches while also revealing new challenges posed by complex multi-DoF joints and intricate kinematic relationships.
We hope Artiverse will serve as a valuable resource for future research in functional 3D modeling, generative object design, and physically grounded reasoning, ultimately enabling richer benchmarks and inspiring new directions in articulable object understanding.
Looking ahead, we plan to steadily grow Artiverse in scale, diversity, and physical realism to support an ever wider range of applications.

%% file: sec/7_suppl.tex
\clearpage
\setcounter{page}{1}
\maketitlesupplementary

\appendix

\section{Additional Statistics}
\label{supp:sec:stats}
In this section, we provide additional analysis of the data.

\subsection{Kinematic Graph Diversity}
\label{supp:sec:graph_diversity}

Besides the part and joint counts presented in the main paper, another important diversity dimension is the diversity of kinematic graphs. Kinematic graphs can act as a proxy of the data complexity, for the categories that allow for various graphs. 

We start by measuring a number of unique graphs, judging by part connectivity, semantic labels and motion types. As the absolute counts are biased by overall and per category data scale, we opt to compute the probability of the occurrence of each graph and measure the entropy of this distribution.

We proceed to compute the effective number (perplexity) of the graphs in the Artiverse and PartNet-Mobility distributions. We find that the effective number of kinematic graphs in Artiverse is 1.5 times higher. Overall, besides the categories that limit the graph diversity (i.e., scissors), Artiverse features more complex part arrangements.

\subsection{Model Selection}
Our dataset construction begins by sourcing high-quality, articulatable models across diverse categories.
We collect and filter models from multiple repositories through a hybrid retrieval workflow that combines automated feature analysis with human oversight.
For well-structured datasets with fine-grained taxonomy, such as HSSD~\cite{khanna2024habitat} and 3DComPaT200~\cite{ahmed20253dcompat200}, we directly remap the objects at the category level.
For large repositories with noisy or inconsistent metadata, we adopt a retrieval-by-example strategy.
We first manually select a small set of models for each sub-category with the help of metadata (e.g., data tags, descriptions, category).
Then we run k-means algorithm on DuoDuoCLIP~\cite{lee2025duoduo} embeddings to find representative \textit{anchor} models that serve as reference exemplars to expand more models in a semi-automated way.
These anchors serve as visual exemplars that guide the retrieval of similar shapes from the model pool.
We render multi-view turntable images for every candidate model and extract visual embeddings using DuoDuoCLIP. 
Similarity scores between candidates and anchors are computed to rank potential matches. 
To ensure category correctness and asset quality, we employ a custom web interface for human-in-the-loop verification of the top-$k$ retrieved candidates.
Models that align with the intended sub-category and exhibit realistic geometry and textures are retained.
The resulting curated collection forms a high-quality starting pool for the downstream annotation pipeline. 

\section{Semi-automated Annotation Pipeline}
\label{supp:sec:pipeline}
In this section, we supply more implementation details of different steps in our data collection and annotation pipeline.

\subsection{Data Preprocessing}
After category-level selection, we preprocess all models to ensure normalization, consistent orientation, metric scale, and geometric uniqueness across the dataset.

To standardize orientation, we convert all assets into a shared y-up coordinate frame and make them zero-centered.
Most objects naturally follow a consistent front-facing direction due to common modeling conventions.
For the remaining minority, we apply axis-aligned rotations to align the canonical front-facing direction.

For metric scaling, we preserve the physical dimensions provided by the original data sources whenever available.
When scale information is missing, we infer a plausible size by sampling from a range of physical dimensions per sub-category.
These ranges are obtained using GPT-4o reasoning and then validated against the objects whose physical sizes are known, ensuring the predicted ranges remain realistic and category-appropriate.

To remove redundant models, we perform near-duplicate detection using DuoduoCLIP~\cite{lee2025duoduo} feature similarity. 
We manually assess models with cosine similarity greater than 0.99 in the specialized interface.
Additionally, we exclude models already present in PartNet-Mobility~\cite{xiang2020sapien} to prevent overlap with existing articulated datasets.

\subsection{Part Segmentation Proposal}
Taking the selected and preprocessed models from previous steps, we annotate functional parts guided by the per-category template (see an example template design in \Cref{fig:template}). 
Our method combines few-shot multi-view segmentation with geometry processing algorithms to produce high-quality part instance proposals that are subsequently verified by human annotators.
The key steps are shown in \Cref{fig:seg_pipeline}.

\mypara{Few-shot semantic segmentation.}
For each category or group of similar categories (e.g., furniture), we manually annotate 5-15 representative objects with functional parts labels.
Using these exemplars, we train the few-shot adaptive segmentation model ASIA~\cite{perla2025asia} on 16 rendered views per object.
Each input image is overlaid with projected 3D part masks sorted by visibility, enabling the model to learn category-specific functional semantics.
Once trained, ASIA is applied to the same 16 views for all objects within the category scope to obtain dense 2D semantic predictions.

\mypara{Propagating 2D semantic regions to 3D.}
We precompute mesh over-segmentation based on triangle connectivity, and project 2D semantic predictions back to these segments.
For each view, we iterate over visible segments and record the semantic labels and their pixel counts.
Each segment is assigned the semantic label with the highest aggregate pixel support across views.
For the occluded segments that are invisible to all rendered views, we find its nearest labeled segment and inherit the corresponding semantic label.
This proximity-based assignment ensures that interior or shadowed regions receive consistent labels without requiring exhaustive rendering.

\mypara{Clustering segments into part instances.}
To derive instance-level parts, we cluster the over-segments within a subset for each semantic label, based on the distance between them.
For each semantic label, we consider all pairs of segments within that label and compute the shortest triangle-triangle distance between them.
These pairwise checks are highly parallelizable.
Segments are considered connected if their minimum distance is below a threshold (e.g., 1 mm)
We then apply a union-find algorithm to obtain disjoint connected subsets, each corresponding to a functional part instance.

\input{figures/part_seg_pipeline.tex}

\subsection{Articulation Proposal}
To generate articulated motion proposals, we design a set of geometric heuristics that infer joint parameters directly from 3D geometry. 
Our goal is to build a general, reusable framework that works across diverse object categories while requiring minimal per-class specialization.

\mypara{Motion type.}
In many cases, there is a single motion type for a particular semantic part. In the case of multiple possible motion types, as determined by the templates, we either use geometric heuristics to disambiguate the case if possible or simply go for the one we expect to be the most commonly present in the distribution. 

\input{figures/contact_region}
\mypara{Motion axis.}
For motion axis, we derive its origin and direction by either relying on the coarse geometric cues (e.g, part centroid, bounding boxes) or from the fine-grained contact region between two kinematically dependent parts, depending on the motion and part type.

For some articulated components (e.g., doors, drawers, buttons), the motion axis is naturally aligned with a characteristic dimension or face of the part bounding box.
To robustly capture these cues, we compute three types of bounding boxes:
1) \textit{AABB} – axis-aligned bounding box obtained from minimum and maximum points of the part geometries;
2) \textit{POBB} – PCA-oriented bounding box whose axes follow principal directions;
3) \textit{GOBB} – gravity-aligned bounding box where the vertical axis fixed to gravity direction and horizontal axes from PCA on the part projected to the ground plane.
We select the bounding box with the smallest volume as the primary geometric descriptor, except when the GOBB volume is within 5\% of the minimum.
In that case, we prefer GOBB because it typically aligns better with human-centric object orientation conventions.

Certain joints require finer geometric reasoning, e.g., rotational joints located off-center (e.g., chair casters), or hinges defined by narrow attachment regions.
For these cases, we identify the contact region between a part and its parent by sampling points from each part surface, computing pairwise distances between the two point sets, and take the nearest points as the contact region.
\Cref{fig:contact} shows some examples of the contact region we detected and used for motion inference across different categories.
If multiple contact regions are detected (e.g., two hinge mounts), the motion axis is inferred from the geometric relationship between these regions, for instance, by connecting region centroids.
This strategy allows reliable axis placement even for small, highly localized joints.

\mypara{Motion range estimation.}
Parts vary in pose and may be partially colliding due to modeling artifacts, making range inference non-trivial.
We therefore simulate joint motion and monitor collision changes with some tolerance to identify the joint limit at two ends.

For rotational joints, we incrementally rotate the part in both directions. 
At each step, we record the number of collision points.
A valid collision event is detected when the count shows a sustained ramp-up over multiple steps and exceeds the initial collision count by a threshold factor.
We take the two collision bounds and derive a safe operational subrange (e.g., 80\% of collision-to-collision span).

For translational joints, we first detect the detachment distance by translating the part until collisions disappear.
Then we sweep in the opposite direction to find the first collision boundary.
These detachment-collision distances are post-processed with category rules to match realistic physical limits.

\mypara{Kinematic graph construction.}
After estimating motion parameters, we construct the kinematic graph describing parent-child relationships between articulated parts.
For each part, we compute pairwise distances between sampled surface points and assign the closest valid parent according to the template constraints.
This results in a complete and consistent kinematic hierarchy that reflects the articulation structure of the object.

\subsection{Interior Completion}
Many static 3D repositories provide only the outward appearance of objects, leaving internal structures either partially modeled or entirely missing.
To ensure functional completeness and realistic physical simulation, we perform interior completion for categories where internal geometry is essential for the following three scenarios.

\mypara{Partial articulated parts.}
Some articulated components are modeled only partially in static datasets.
A common example is a drawer, where the front panel exists but the drawer box is missing—leading to unrealistic behavior once articulation is introduced.
For such cases, we complete the missing geometry using a heuristic adapted from prior work S2O~\cite{iliash2024s2o}: 
we cast rays behind the drawer's front panel to infer the enclosed volume and procedurally insert side panels, a bottom panel, and a back panel consistent with the enclosing cavity.

\mypara{Missing articulated parts.}
Large appliances often rely on internal movable components that are not modeled in the original mesh but are essential for functional interactivity, e.g., baskets in dishwashers, turntables in microwaves, or containers in refrigerators.
For categories requiring such internal articulated elements, we collect set of representative instances from the existing assets to define the expected interior parts.
Next, when processing other objects sharing the same category, we detect whether such elements already exist from upstream segmentation; if not, we procedurally insert interior articulated components by randomly retrieving geometry templates from annotated exemplars or synthesizing simple parametric versions aligned with the internal cavity.

\mypara{Missing interaction affordance parts.}
Many furniture and storage objects lack internal shelving or structural dividers, leading to hollow cavities that break realism and limit utility for more interactivity.
For targeted categories (e.g., cabinets, wardrobes, refrigerators), we check segmentation results to determine whether interior components are present.
If the interior is empty, we procedurally generate shelves, racks, or rails based on the object's internal bounding volume.
These elements are aligned with the geometry of the enclosing body and follow category-specific placement patterns (e.g., horizontal shelves for wardrobes, horizontal racks for armoires, additional vertical dividers for cabinets).

This structured interior completion pipeline ensures that objects are not only visually consistent but also functionally complete, enabling realistic articulated behavior and physically plausible interaction in downstream simulations.

\subsection{Manual verification details}
Each data sample undergoes a total of 4 verification stages. Namely, segmentation correction, segmentation verification, motion correction, and final verification. We dedicate 2 steps to segmentation as we find it the most prone to human errors. Correction is performed in the 3D annotation tool directly, while verification is performed in specialized UI where the expert annotators are being presented with a video for each part, with the part of interest highlighted and other parts semi-transparent, with the camera revolving around that part. Motion correction is also performed in 3D motion annotation tool. In the final verification, annotators are presented with the videos of each single part moving, both highlighted and textured. These objects also include the completed interiors, and material proposals, which users verify in this stage only. 

The recruited annotators are mostly grad and undergrad students recruited within the institution. Throughout all the annotation sessions, a total of 33 annotators have contributed. 18 of them, contributing more than 5 hours total. The overall annotation time across all the manual correction and verification steps is 400 hours. 

\subsection{Pipeline limitations}
Most of the failure modes come from the geometry artifacts, such as: thin structures, duplicated surfaces, parts that are not modeled with separate connected components. Another limitation of the pipeline appears in the cases of complex kinematic chains such as folding chairs. In these cases, we adopt simplified kinematic model which might ignore or remove some parts. Finally, our current representation does not allow modeling of deformable parts, which otherwise would contribute to the physical realism of the data.

\section{Experiments}
In this section we provide additional implementation details of the experiments and provide more results.

\subsection{Implementation details}
For FPNGroupMot, we first pre-process our data to merge even the interactable parts, such as handles, into the articulated parts. We extend the method with an extra axis prediction head that allows support for the universal joint. The metrics follow the same setup as in S2O~\cite{iliash2024s2o}, except we ignore semantic labels due to the mismatch between PM and Artiverse while aiming at cross-validation. We increase the batch size to 32 and the number of training epochs to 600, otherwise keeping the hyperparameters in accordance to the original paper. Training is conducted on 4 NVIDIA L40S GPUs for 20 hours on Artiverse and 13 hours on PM. 

Due to the flaw of the original design of label representation, SINGAPO cannot be scaled well to the large number of parts. This fact significantly impacts the performance as semantic labels are used for retrieval. In order to setup a fair comparison, we enhance augment SINGAPO with generative models with strong image-to-3D prior at inference time. In particular, the image is first passed through TRELLIS~\cite{xiang2025structured} to obtain the initial 3D geometry. This geometry, combined with SINGAPO bounding boxes is then used to condition X-Part~\cite{yan2025xpart} for per-part generation. Furthermore, we extend SINGAPO with an extra axis and range attributes to support universal joints. We increase the batch size to 120 and perform training on 2 NVIDIA H200 GPUs for 18 hours.

\subsection{Image to Articulated object generation}
The qualitative comparison on image to articulated object task is presented in \Cref{fig:singapo_aa_qualitative}. We find that Articulate Anything is heavily bottlenecked by retrieval, which is especially problematic on our diverse data. While SINGAPO combined with geometry generation excels both at structural understanding and geometry reconstruction. However, we find that SINGAPO is limited by occasional misalignment of TRELLIS geometry with SINGAPO bounding boxes that are used to condition XPart.

\input{figures/singapo_aa_qualitative}

\input{figures/example_template}
\input{figures/thumbnails_figure}

%% file: figures/part_seg_pipeline.tex
\begin{figure}[h]
\centering
\includegraphics[width=\linewidth]{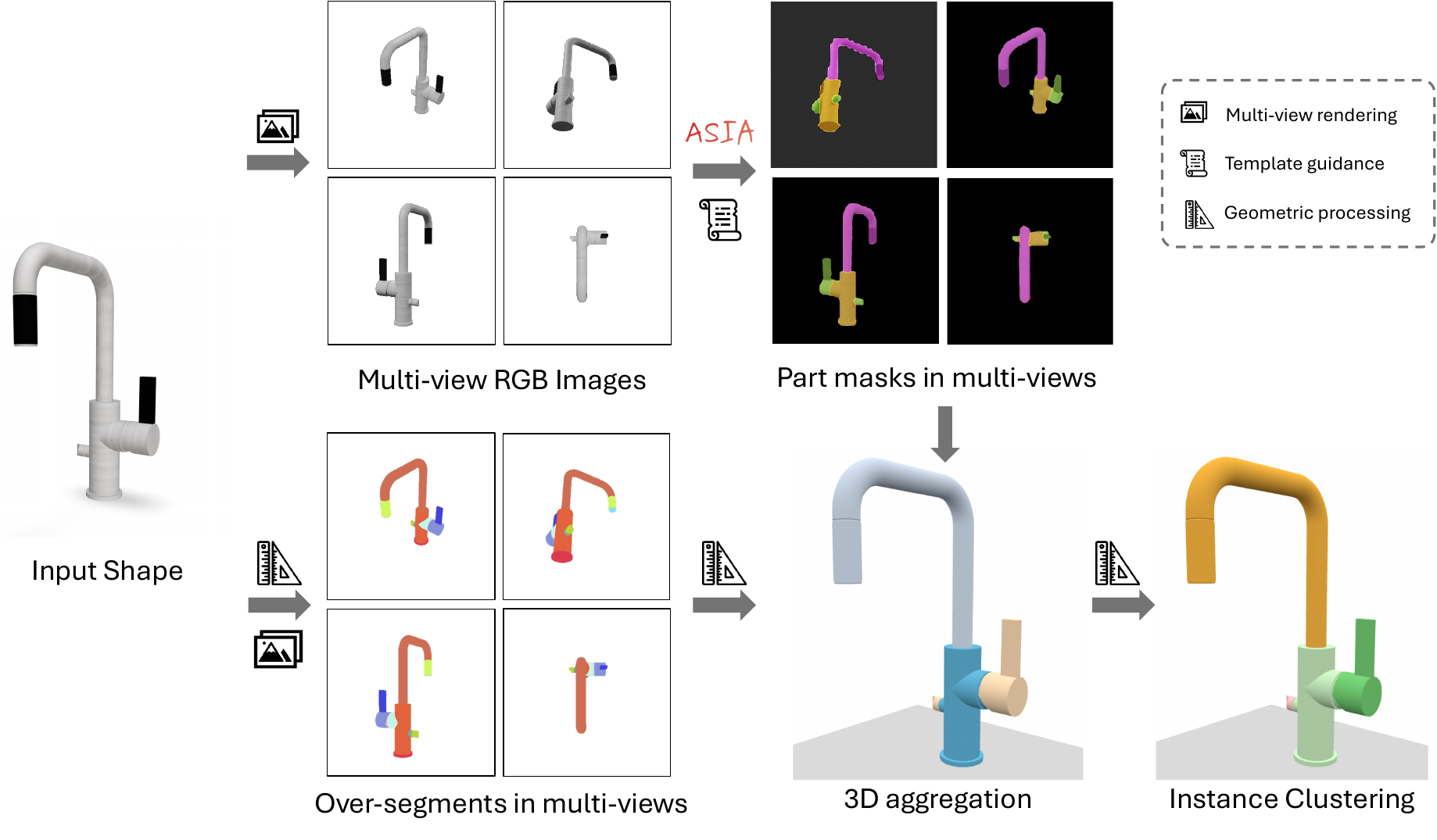}
\caption{
Key steps of proposing functional part segmentation. 
Given an input mesh, we
1) few-shot an adaptive segmentation model to predict masks on multi-view images for the parts predefined in the template;
2) propagate 2D semantic regions from multi-views to over-segments by assigning labels to mesh over-segments, including occluded regions, through visibility voting and proximity reasoning; and
3) cluster part instances into final part instances based on point-wise distanceb between over-segments.
} 
\label{fig:seg_pipeline}
\end{figure}

%% file: figures/contact_region.tex
\begin{figure}[t]
\centering
\includegraphics[width=\linewidth]{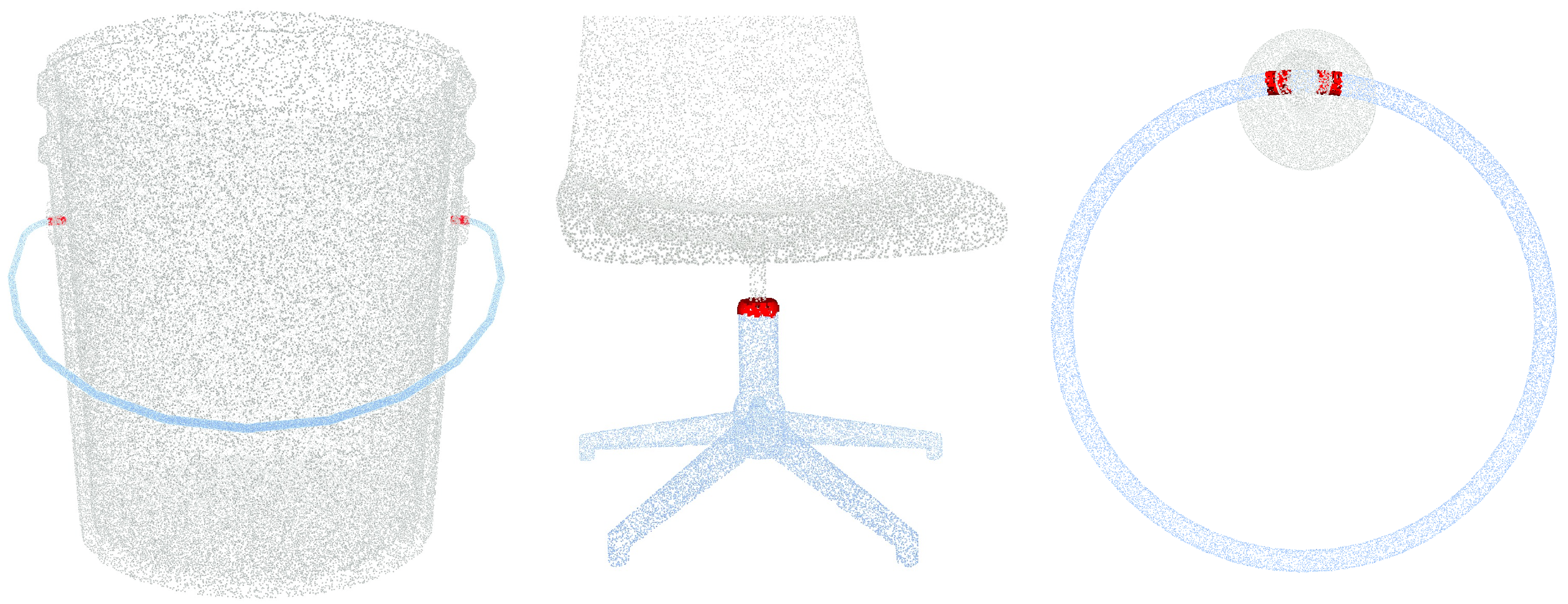}
\caption{
Illustration of the contact region (highlighed in red points) we detected between two kinematically dependent parts (colored in grey and blue) to infer the motion joints.
} 
\label{fig:contact}
\end{figure}

%% file: figures/singapo_aa_qualitative.tex
\begin{figure}[ht]
\centering

\renewcommand{\arraystretch}{1.4}

\begin{tabular}{ccc}
Input & AA & SINGAPO \\

\hline

\includegraphics[trim=50 50 50 50, clip,width=0.22\linewidth]{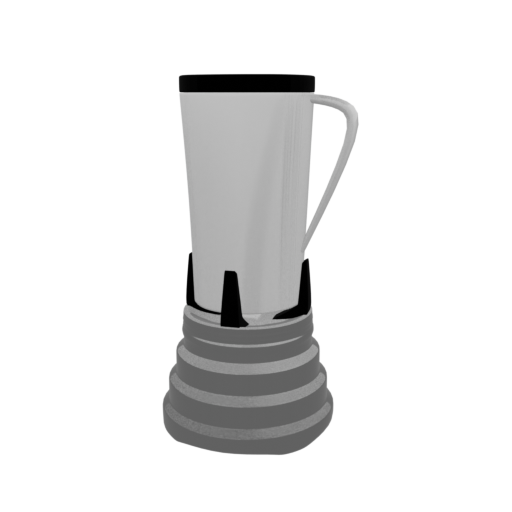} &
\includegraphics[trim=200 200 200 200, clip, width=0.22\linewidth]{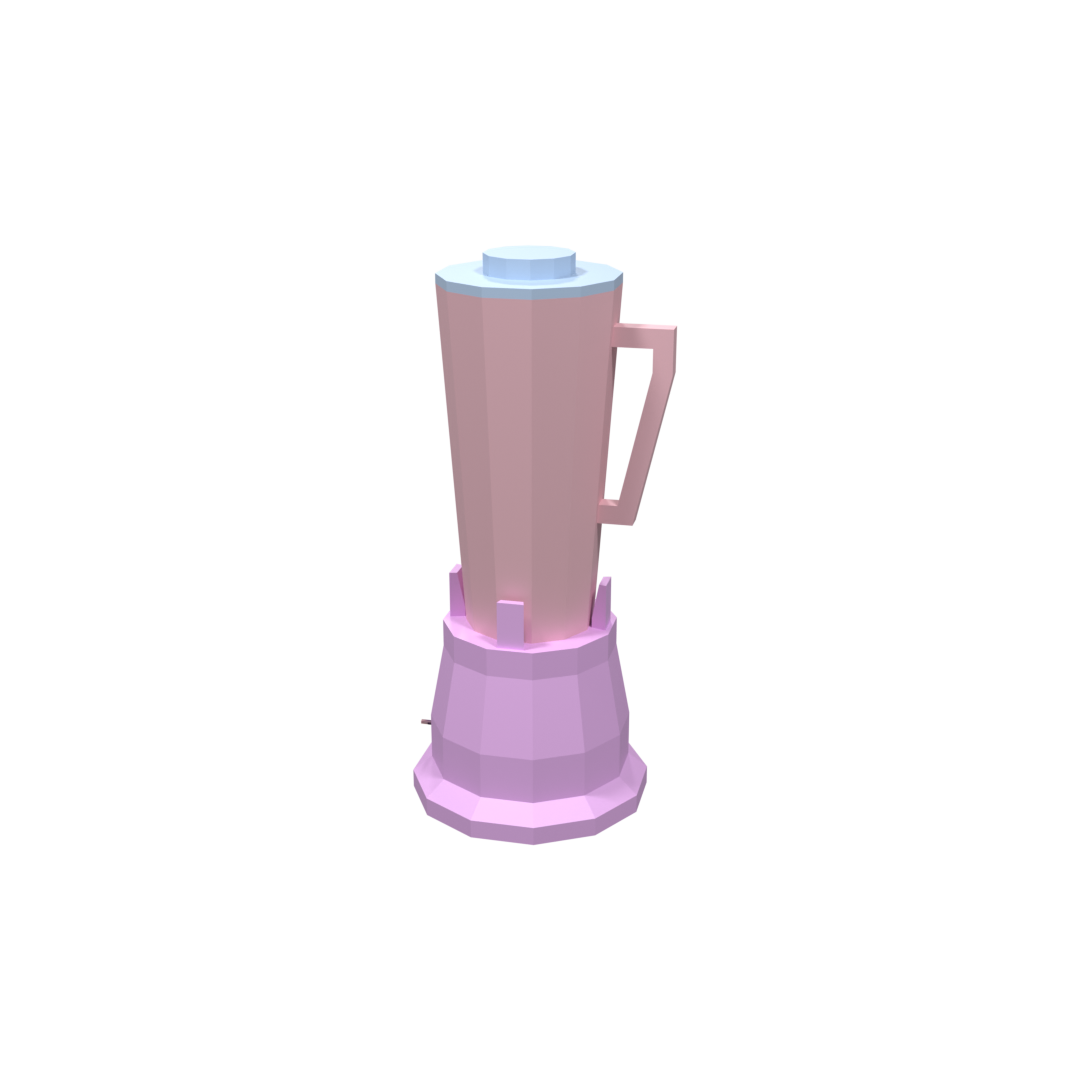} &
\includegraphics[trim=130 130 130 130, clip,width=0.22\linewidth]{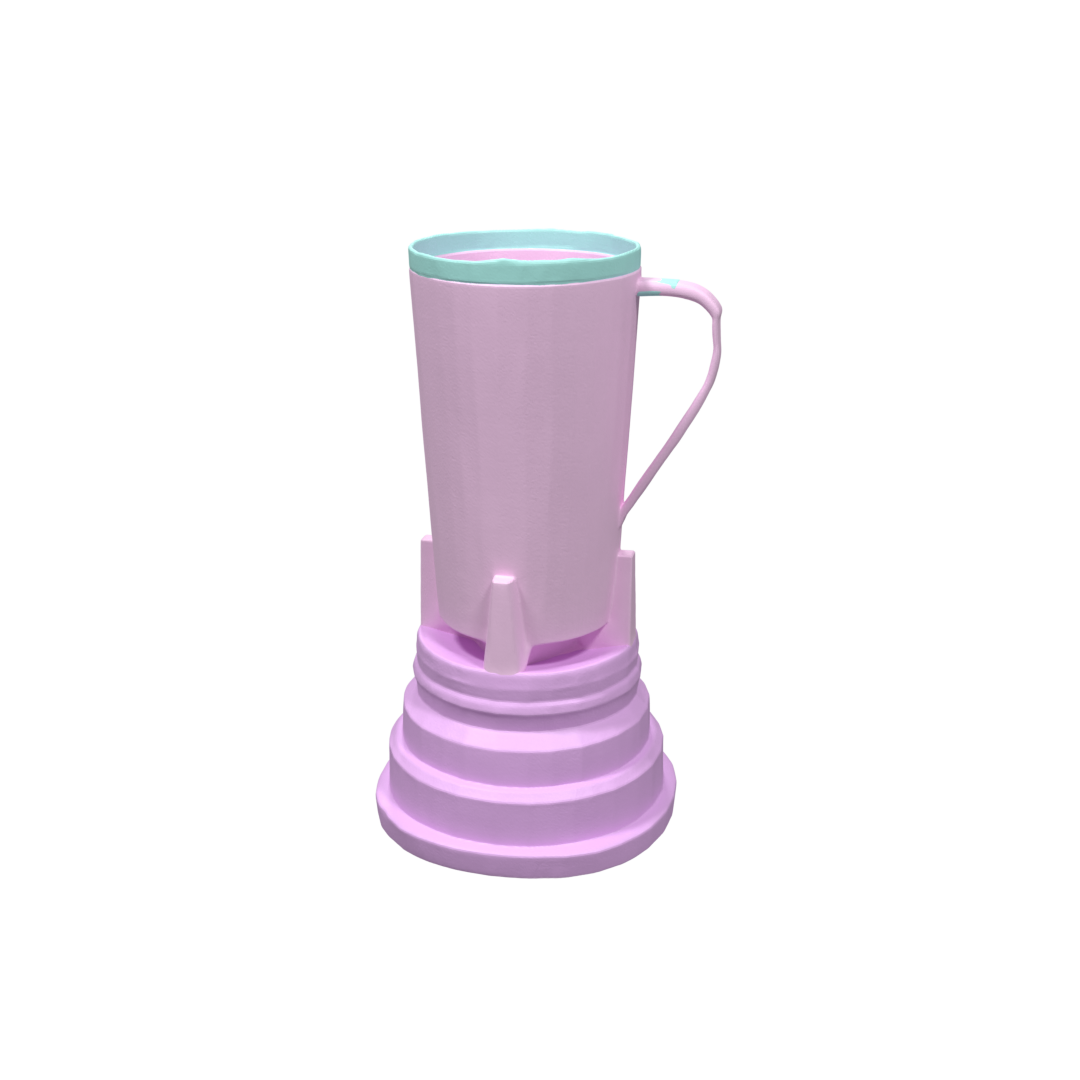} \\

\includegraphics[trim=50 50 50 70, clip,  width=0.22\linewidth]{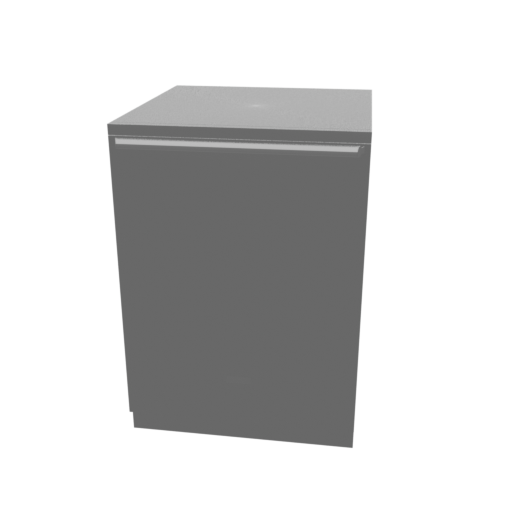} &
\includegraphics[trim=150 150 150 170, clip, width=0.22\linewidth]{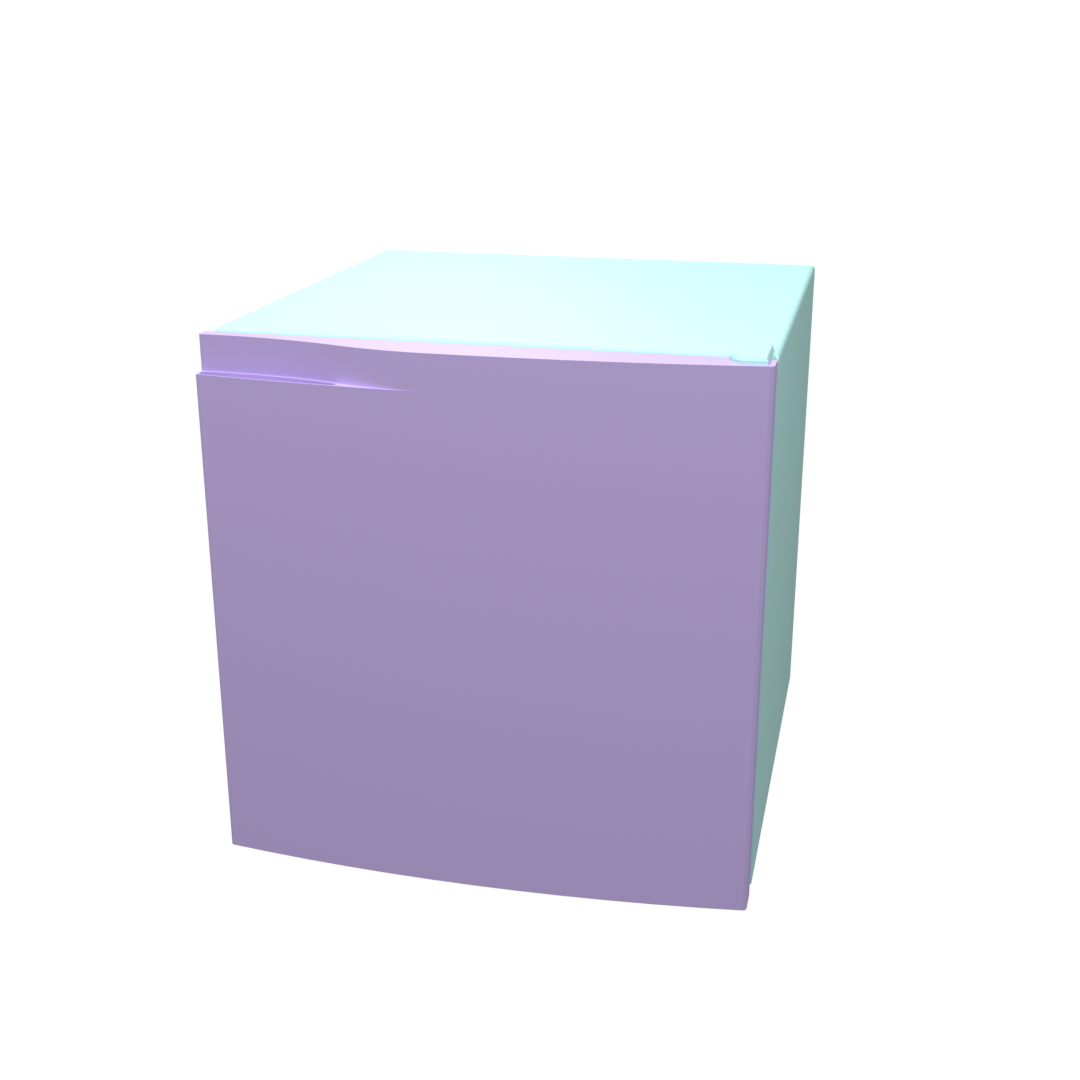} &
\includegraphics[trim=130 130 130 150, clip, width=0.22\linewidth]{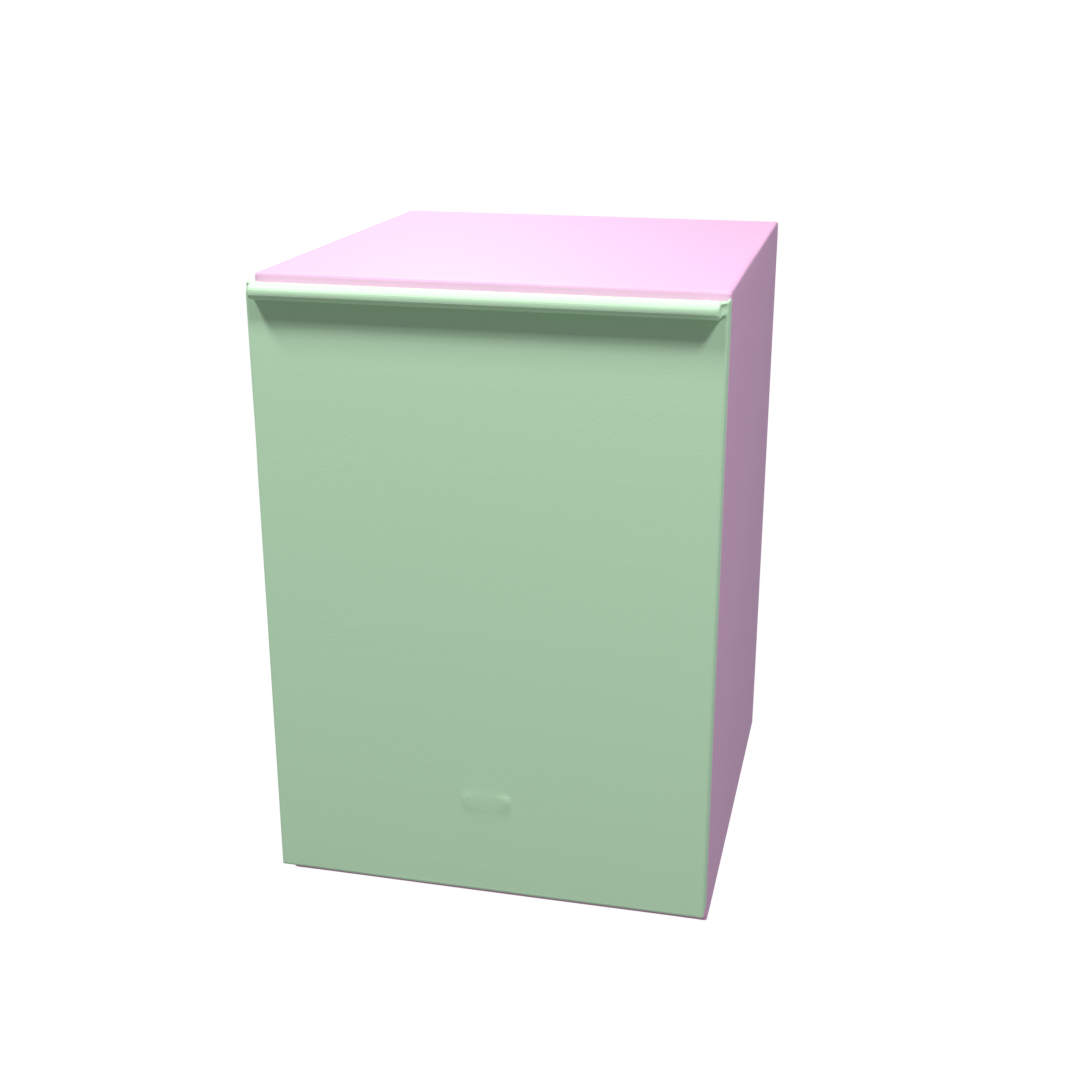} \\

\includegraphics[trim=70 50 70 70, clip, width=0.22\linewidth]{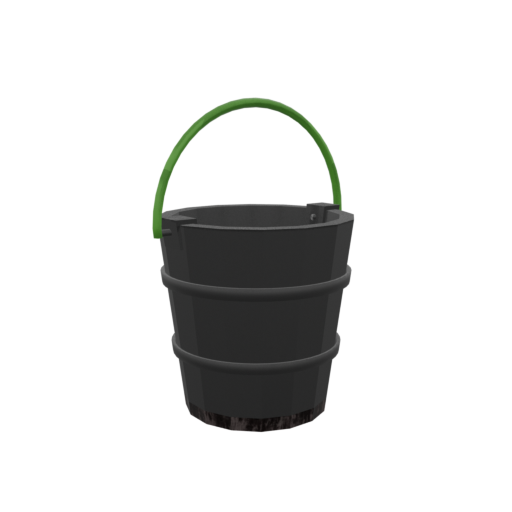} &
\includegraphics[trim=200 250 200 200, clip,width=0.22\linewidth]{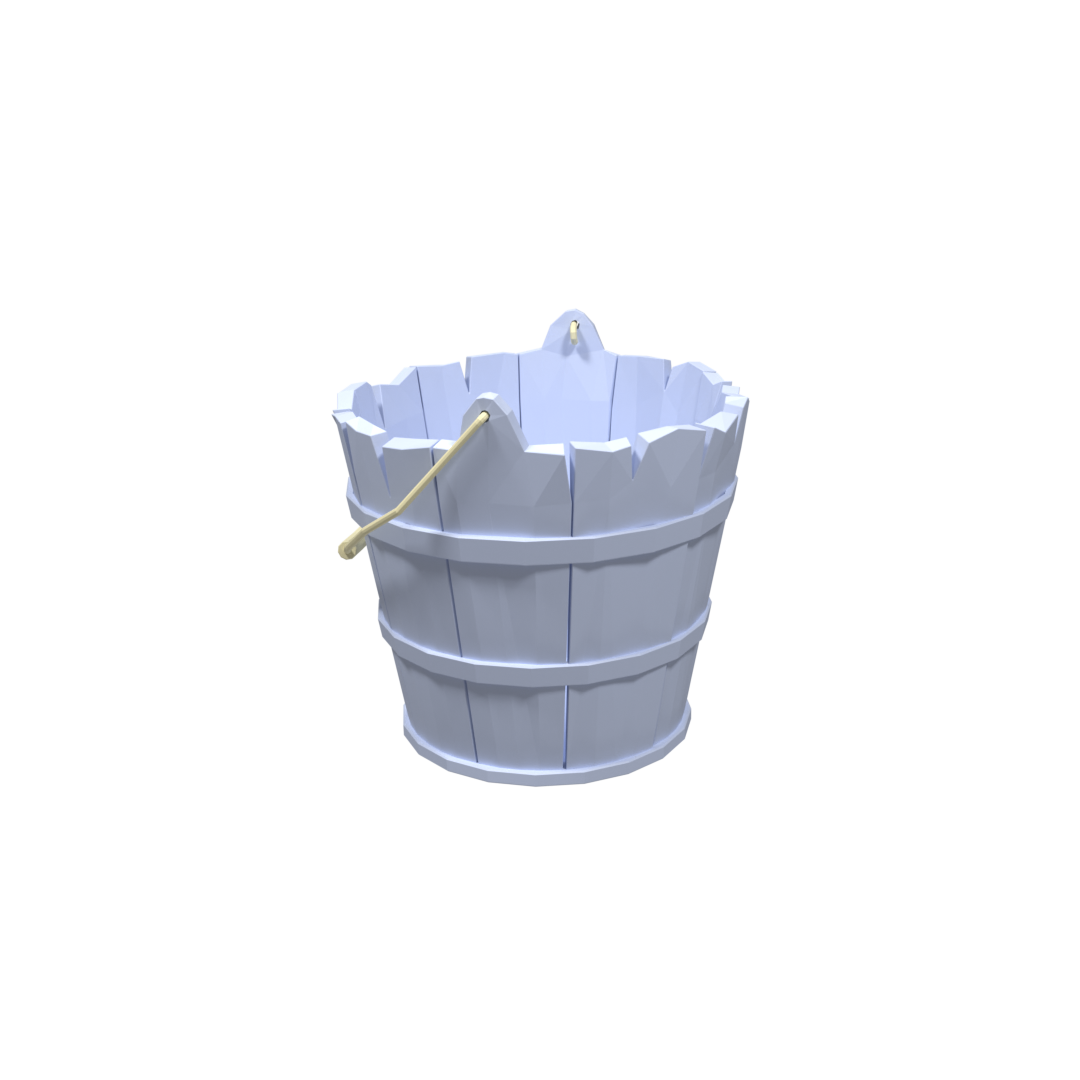} &
\includegraphics[trim=200 180 200 200, clip,width=0.22\linewidth]{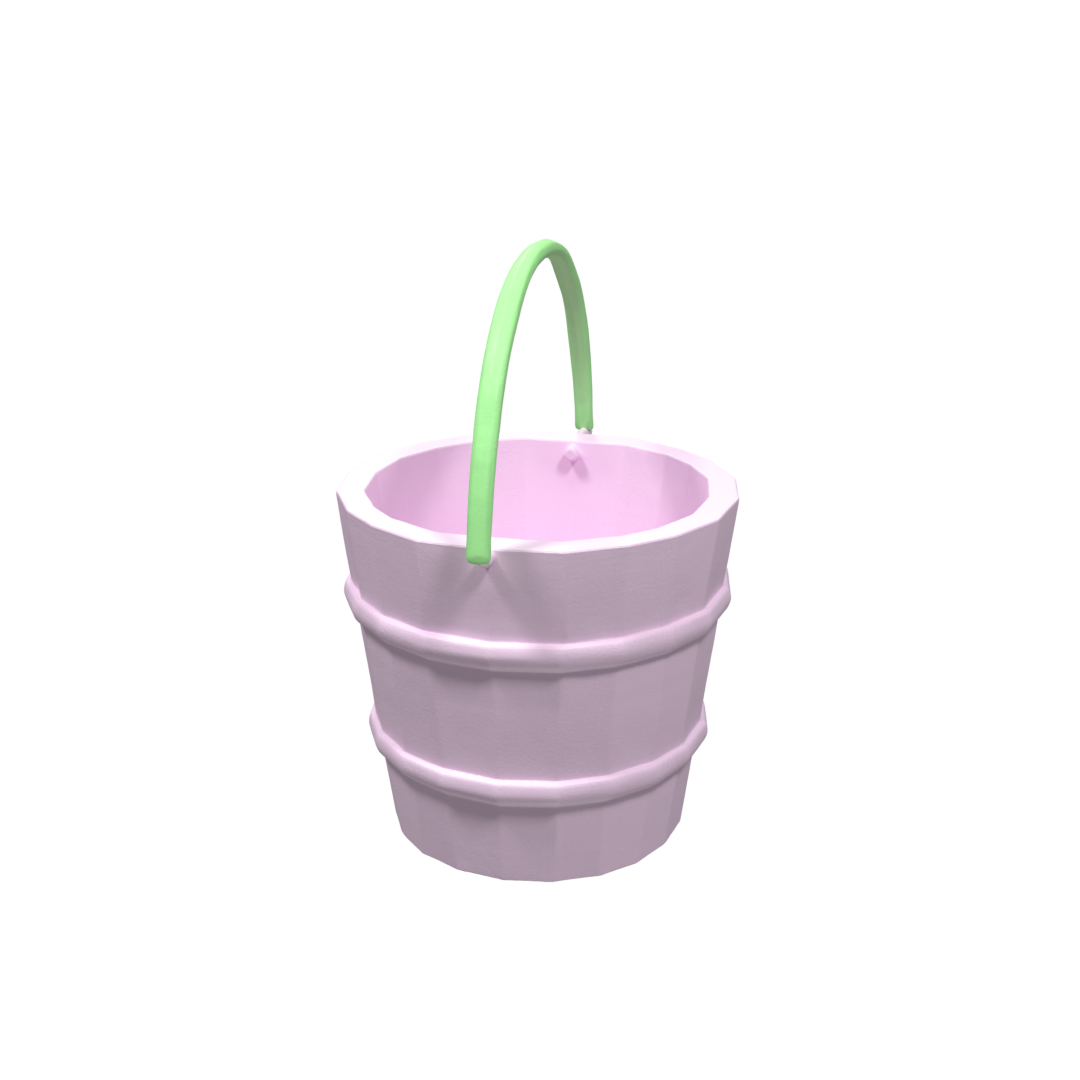} \\

\includegraphics[trim=50 50 50 50, clip,width=0.22\linewidth]{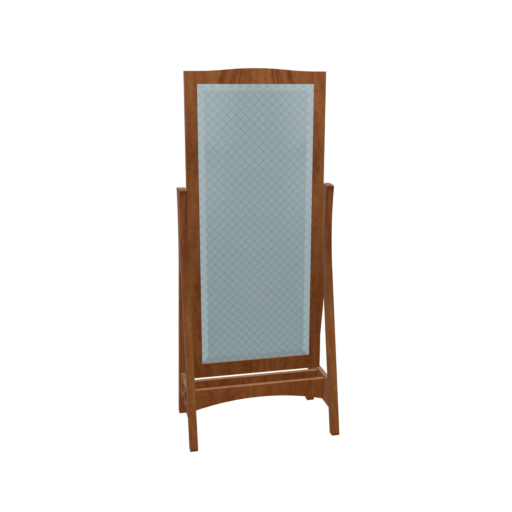} &
\includegraphics[trim=80 80 80 80, clip,width=0.22\linewidth]{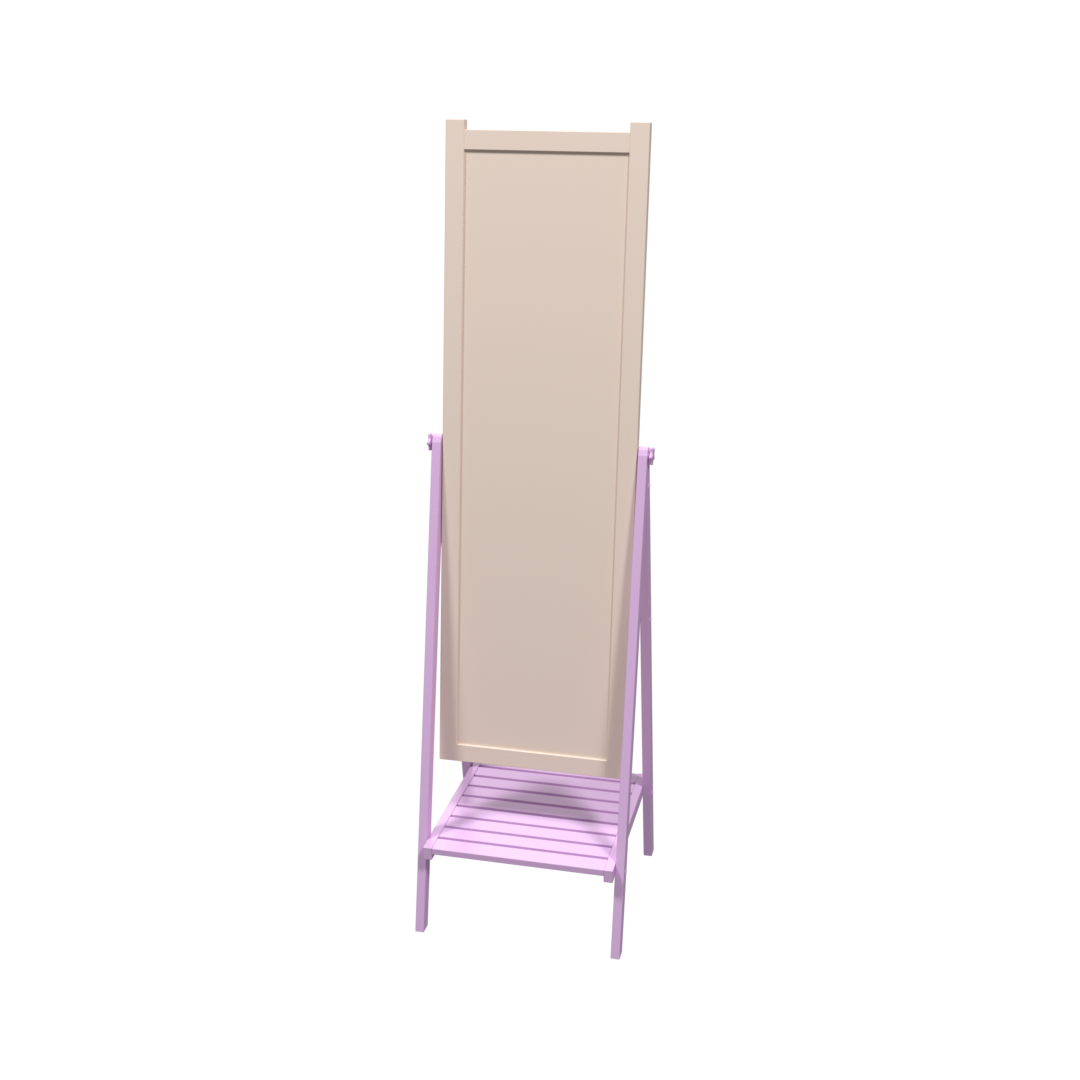} &
\includegraphics[trim=200 200 200 200, clip, width=0.22\linewidth]{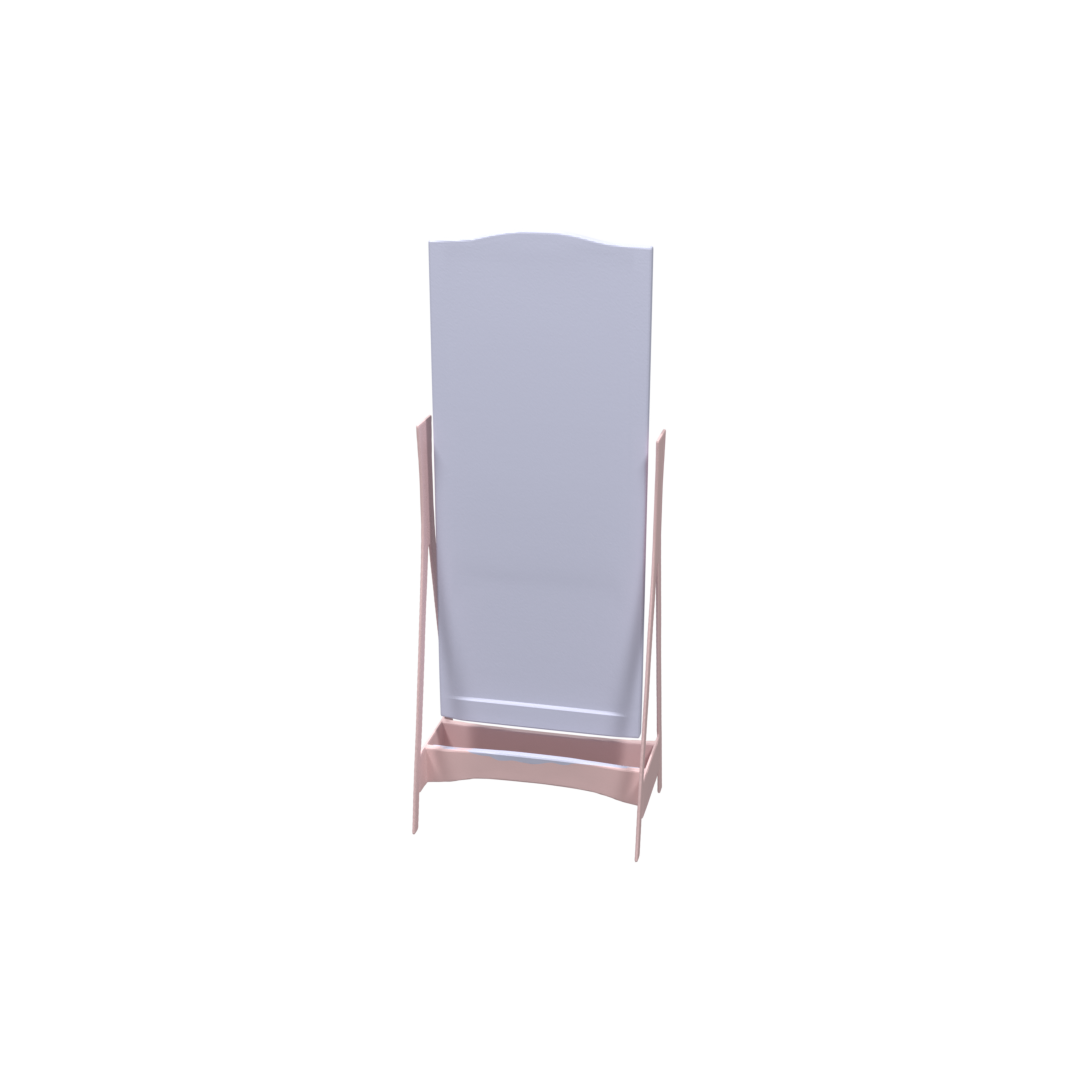} \\

\end{tabular}

\caption{Qualitative results of Articulate Anything compared to SINGAPO + XPart on Artiverse. Articulate Anything is limited by retrieval quality while SINGAPO combined with XPart allows adapting to the input geometry, while predicting reliable object structures.}
\label{fig:singapo_aa_qualitative}
\end{figure}

%% file: figures/example_template.tex
\begin{figure*}[h]
\centering
\begin{minipage}{0.95\linewidth}
\begin{lstlisting}[language=json]
{
    "main_category": "fan",
    "gloss": "A household electrical appliance that produces airflow ...",
    "content": [
      {
        "name": "base",
        "gloss": "The bottom support structure of the fan that provides stability and houses electrical connections.",
        "kinematic": {
          "articulatable": false,
          "link_dependency": [],
          "joint_type": ["fixed"]
        }
      },
      {
        "name": "pole",
        "gloss": "The vertical support column connecting the base to the head ...",
        "kinematic": {
          "articulatable": true,
          "link_dependency": ["base"],
          "joint_type": ["cylindrical", "fixed", "revolute", "prismatic"]
        }
      },
      {
        "name": "head",
        "gloss": "The main assembly containing the motor and blades, mounted on the pole ...",
        "kinematic": {
          "articulatable": true,
          "link_dependency": ["pole"],
          "joint_type": ["revolute", "fixed", "universal"]
        }
      },
      {
        "name": "blade_set",
        "gloss": "A set of rotating aerofoil components attached to the motor shaft ...",
        "parts": [
          {
            "name": "blade",
            "gloss": "..."
          },
          {
            "name": "rotor",
            "gloss": "..."
          }
        ],
        "kinematic": {
          "articulatable": true,
          "link_dependency": ["head"],
          "joint_type": ["continuous"]
        }
      },
      ...
    ]
  }
  
\end{lstlisting}
\end{minipage}
\caption{An example of our designed template. }
\label{fig:template}
\end{figure*}

%% file: figures/thumbnails_figure.tex

\onecolumn
\begin{figure}[p]
\centering
{\footnotesize\setlength{\tabcolsep}{1pt}
\begin{tikzpicture}
  \node[inner sep=0] (img) {\includegraphics[width=0.115\textwidth]{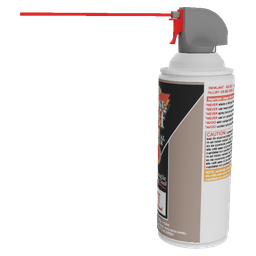}};
  \node[anchor=south east,font=\scriptsize,fill=white,fill opacity=0.8,text opacity=1,inner sep=2pt]
      at (img.south east) {aerosol can};
\end{tikzpicture} 
 \begin{tikzpicture}
  \node[inner sep=0] (img) {\includegraphics[width=0.115\textwidth]{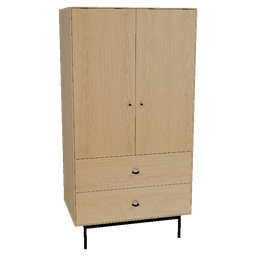}};
  \node[anchor=south east,font=\scriptsize,fill=white,fill opacity=0.8,text opacity=1,inner sep=2pt]
      at (img.south east) {armoire};
\end{tikzpicture} 
 \begin{tikzpicture}
  \node[inner sep=0] (img) {\includegraphics[width=0.115\textwidth]{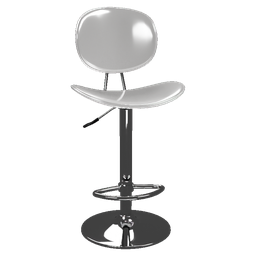}};
  \node[anchor=south east,font=\scriptsize,fill=white,fill opacity=0.8,text opacity=1,inner sep=2pt]
      at (img.south east) {bar stool};
\end{tikzpicture} 
 \begin{tikzpicture}
  \node[inner sep=0] (img) {\includegraphics[width=0.115\textwidth]{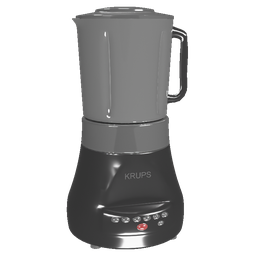}};
  \node[anchor=south east,font=\scriptsize,fill=white,fill opacity=0.8,text opacity=1,inner sep=2pt]
      at (img.south east) {blender};
\end{tikzpicture} 
 \begin{tikzpicture}
  \node[inner sep=0] (img) {\includegraphics[width=0.115\textwidth]{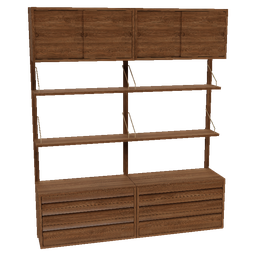}};
  \node[anchor=south east,font=\scriptsize,fill=white,fill opacity=0.8,text opacity=1,inner sep=2pt]
      at (img.south east) {bookcase};
\end{tikzpicture} 
 \begin{tikzpicture}
  \node[inner sep=0] (img) {\includegraphics[width=0.115\textwidth]{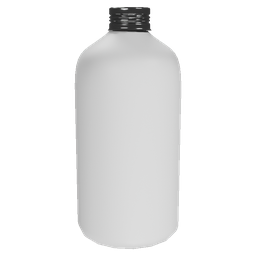}};
  \node[anchor=south east,font=\scriptsize,fill=white,fill opacity=0.8,text opacity=1,inner sep=2pt]
      at (img.south east) {bottle};
\end{tikzpicture} 
 \begin{tikzpicture}
  \node[inner sep=0] (img) {\includegraphics[width=0.115\textwidth]{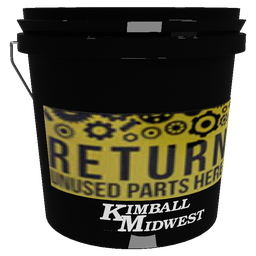}};
  \node[anchor=south east,font=\scriptsize,fill=white,fill opacity=0.8,text opacity=1,inner sep=2pt]
      at (img.south east) {bucket};
\end{tikzpicture} 
 \begin{tikzpicture}
  \node[inner sep=0] (img) {\includegraphics[width=0.115\textwidth]{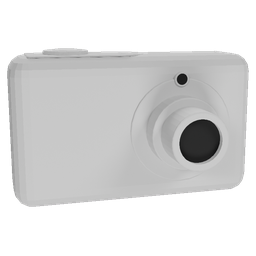}};
  \node[anchor=south east,font=\scriptsize,fill=white,fill opacity=0.8,text opacity=1,inner sep=2pt]
      at (img.south east) {camera};
\end{tikzpicture} \\
\begin{tikzpicture}
  \node[inner sep=0] (img) {\includegraphics[width=0.115\textwidth]{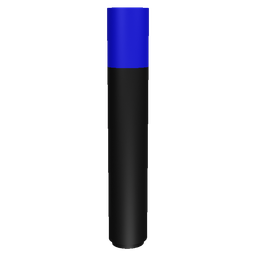}};
  \node[anchor=south east,font=\scriptsize,fill=white,fill opacity=0.8,text opacity=1,inner sep=2pt]
      at (img.south east) {cap pen};
\end{tikzpicture} 
 \begin{tikzpicture}
  \node[inner sep=0] (img) {\includegraphics[width=0.115\textwidth]{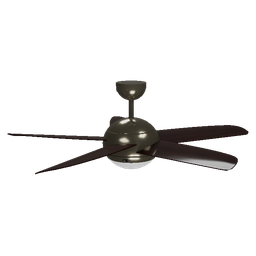}};
  \node[anchor=south east,font=\scriptsize,fill=white,fill opacity=0.8,text opacity=1,inner sep=2pt]
      at (img.south east) {ceiling fan};
\end{tikzpicture} 
 \begin{tikzpicture}
  \node[inner sep=0] (img) {\includegraphics[width=0.115\textwidth]{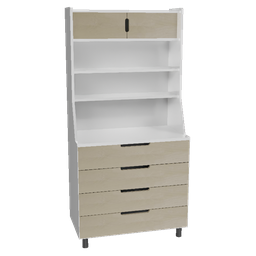}};
  \node[anchor=south east,font=\scriptsize,fill=white,fill opacity=0.8,text opacity=1,inner sep=2pt]
      at (img.south east) {chest of drawers};
\end{tikzpicture} 
 \begin{tikzpicture}
  \node[inner sep=0] (img) {\includegraphics[width=0.115\textwidth]{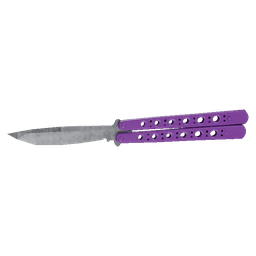}};
  \node[anchor=south east,font=\scriptsize,fill=white,fill opacity=0.8,text opacity=1,inner sep=2pt]
      at (img.south east) {clasp knife};
\end{tikzpicture} 
 \begin{tikzpicture}
  \node[inner sep=0] (img) {\includegraphics[width=0.115\textwidth]{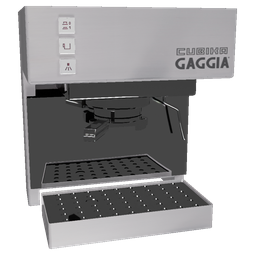}};
  \node[anchor=south east,font=\scriptsize,fill=white,fill opacity=0.8,text opacity=1,inner sep=2pt]
      at (img.south east) {coffee maker};
\end{tikzpicture} 
 \begin{tikzpicture}
  \node[inner sep=0] (img) {\includegraphics[width=0.115\textwidth]{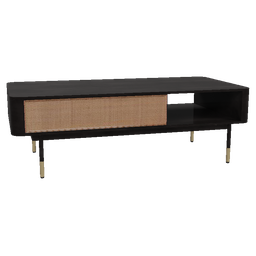}};
  \node[anchor=south east,font=\scriptsize,fill=white,fill opacity=0.8,text opacity=1,inner sep=2pt]
      at (img.south east) {coffee table};
\end{tikzpicture} 
 \begin{tikzpicture}
  \node[inner sep=0] (img) {\includegraphics[width=0.115\textwidth]{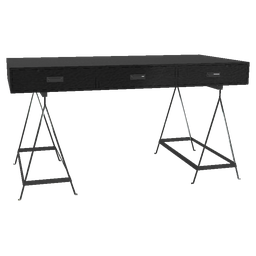}};
  \node[anchor=south east,font=\scriptsize,fill=white,fill opacity=0.8,text opacity=1,inner sep=2pt]
      at (img.south east) {desk};
\end{tikzpicture} 
 \begin{tikzpicture}
  \node[inner sep=0] (img) {\includegraphics[width=0.115\textwidth]{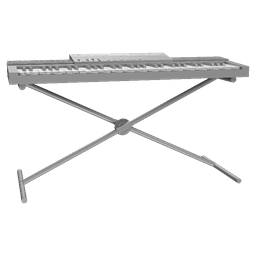}};
  \node[anchor=south east,font=\scriptsize,fill=white,fill opacity=0.8,text opacity=1,inner sep=2pt]
      at (img.south east) {digital piano};
\end{tikzpicture} \\
\begin{tikzpicture}
  \node[inner sep=0] (img) {\includegraphics[width=0.115\textwidth]{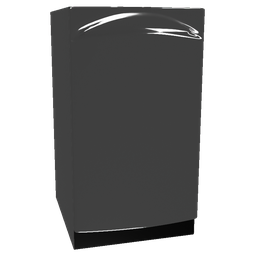}};
  \node[anchor=south east,font=\scriptsize,fill=white,fill opacity=0.8,text opacity=1,inner sep=2pt]
      at (img.south east) {dishwasher};
\end{tikzpicture} 
 \begin{tikzpicture}
  \node[inner sep=0] (img) {\includegraphics[width=0.115\textwidth]{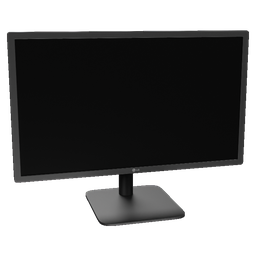}};
  \node[anchor=south east,font=\scriptsize,fill=white,fill opacity=0.8,text opacity=1,inner sep=2pt]
      at (img.south east) {display};
\end{tikzpicture} 
 \begin{tikzpicture}
  \node[inner sep=0] (img) {\includegraphics[width=0.115\textwidth]{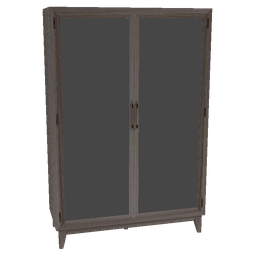}};
  \node[anchor=south east,font=\scriptsize,fill=white,fill opacity=0.8,text opacity=1,inner sep=2pt]
      at (img.south east) {display cabinet};
\end{tikzpicture} 
 \begin{tikzpicture}
  \node[inner sep=0] (img) {\includegraphics[width=0.115\textwidth]{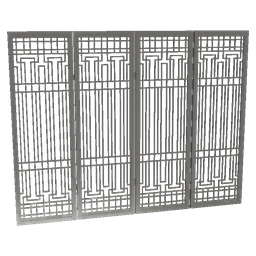}};
  \node[anchor=south east,font=\scriptsize,fill=white,fill opacity=0.8,text opacity=1,inner sep=2pt]
      at (img.south east) {divider};
\end{tikzpicture} 
 \begin{tikzpicture}
  \node[inner sep=0] (img) {\includegraphics[width=0.115\textwidth]{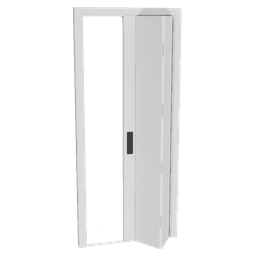}};
  \node[anchor=south east,font=\scriptsize,fill=white,fill opacity=0.8,text opacity=1,inner sep=2pt]
      at (img.south east) {door};
\end{tikzpicture} 
 \begin{tikzpicture}
  \node[inner sep=0] (img) {\includegraphics[width=0.115\textwidth]{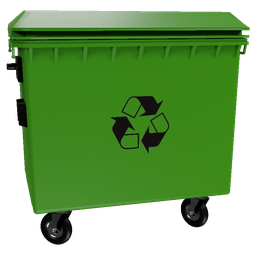}};
  \node[anchor=south east,font=\scriptsize,fill=white,fill opacity=0.8,text opacity=1,inner sep=2pt]
      at (img.south east) {dumpster};
\end{tikzpicture} 
 \begin{tikzpicture}
  \node[inner sep=0] (img) {\includegraphics[width=0.115\textwidth]{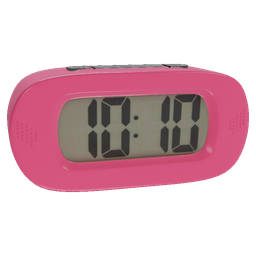}};
  \node[anchor=south east,font=\scriptsize,fill=white,fill opacity=0.8,text opacity=1,inner sep=2pt]
      at (img.south east) {electric clock};
\end{tikzpicture} 
 \begin{tikzpicture}
  \node[inner sep=0] (img) {\includegraphics[width=0.115\textwidth]{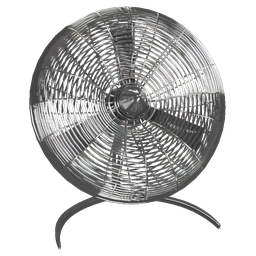}};
  \node[anchor=south east,font=\scriptsize,fill=white,fill opacity=0.8,text opacity=1,inner sep=2pt]
      at (img.south east) {electric fan};
\end{tikzpicture} \\
\begin{tikzpicture}
  \node[inner sep=0] (img) {\includegraphics[width=0.115\textwidth]{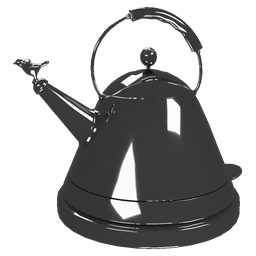}};
  \node[anchor=south east,font=\scriptsize,fill=white,fill opacity=0.8,text opacity=1,inner sep=2pt]
      at (img.south east) {electric kettle};
\end{tikzpicture} 
 \begin{tikzpicture}
  \node[inner sep=0] (img) {\includegraphics[width=0.115\textwidth]{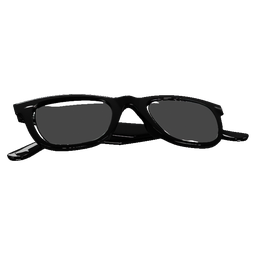}};
  \node[anchor=south east,font=\scriptsize,fill=white,fill opacity=0.8,text opacity=1,inner sep=2pt]
      at (img.south east) {eyeglasses};
\end{tikzpicture} 
 \begin{tikzpicture}
  \node[inner sep=0] (img) {\includegraphics[width=0.115\textwidth]{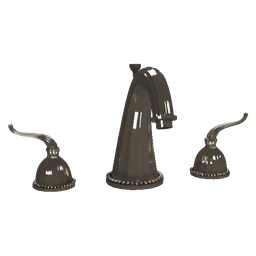}};
  \node[anchor=south east,font=\scriptsize,fill=white,fill opacity=0.8,text opacity=1,inner sep=2pt]
      at (img.south east) {faucet};
\end{tikzpicture} 
 \begin{tikzpicture}
  \node[inner sep=0] (img) {\includegraphics[width=0.115\textwidth]{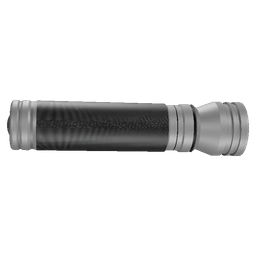}};
  \node[anchor=south east,font=\scriptsize,fill=white,fill opacity=0.8,text opacity=1,inner sep=2pt]
      at (img.south east) {flashlight};
\end{tikzpicture} 
 \begin{tikzpicture}
  \node[inner sep=0] (img) {\includegraphics[width=0.115\textwidth]{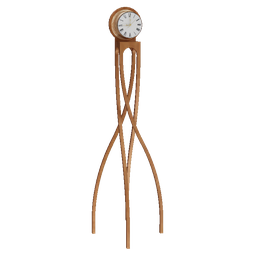}};
  \node[anchor=south east,font=\scriptsize,fill=white,fill opacity=0.8,text opacity=1,inner sep=2pt]
      at (img.south east) {floor clock};
\end{tikzpicture} 
 \begin{tikzpicture}
  \node[inner sep=0] (img) {\includegraphics[width=0.115\textwidth]{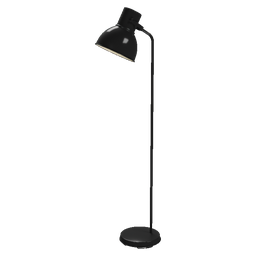}};
  \node[anchor=south east,font=\scriptsize,fill=white,fill opacity=0.8,text opacity=1,inner sep=2pt]
      at (img.south east) {floor lamp};
\end{tikzpicture} 
 \begin{tikzpicture}
  \node[inner sep=0] (img) {\includegraphics[width=0.115\textwidth]{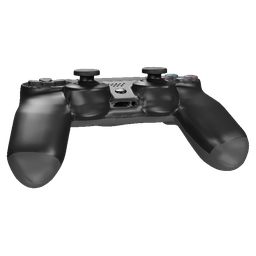}};
  \node[anchor=south east,font=\scriptsize,fill=white,fill opacity=0.8,text opacity=1,inner sep=2pt]
      at (img.south east) {game controller};
\end{tikzpicture} 
 \begin{tikzpicture}
  \node[inner sep=0] (img) {\includegraphics[width=0.115\textwidth]{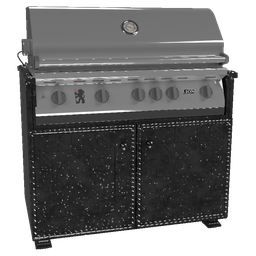}};
  \node[anchor=south east,font=\scriptsize,fill=white,fill opacity=0.8,text opacity=1,inner sep=2pt]
      at (img.south east) {grill};
\end{tikzpicture} \\
\begin{tikzpicture}
  \node[inner sep=0] (img) {\includegraphics[width=0.115\textwidth]{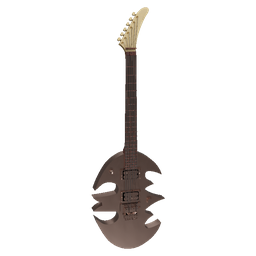}};
  \node[anchor=south east,font=\scriptsize,fill=white,fill opacity=0.8,text opacity=1,inner sep=2pt]
      at (img.south east) {guitar};
\end{tikzpicture} 
 \begin{tikzpicture}
  \node[inner sep=0] (img) {\includegraphics[width=0.115\textwidth]{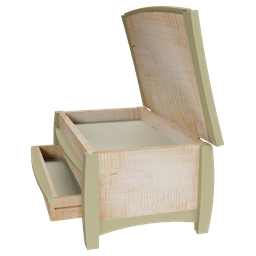}};
  \node[anchor=south east,font=\scriptsize,fill=white,fill opacity=0.8,text opacity=1,inner sep=2pt]
      at (img.south east) {jewelry box};
\end{tikzpicture} 
 \begin{tikzpicture}
  \node[inner sep=0] (img) {\includegraphics[width=0.115\textwidth]{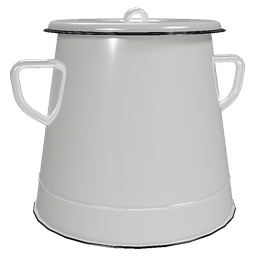}};
  \node[anchor=south east,font=\scriptsize,fill=white,fill opacity=0.8,text opacity=1,inner sep=2pt]
      at (img.south east) {kitchen pot};
\end{tikzpicture} 
 \begin{tikzpicture}
  \node[inner sep=0] (img) {\includegraphics[width=0.115\textwidth]{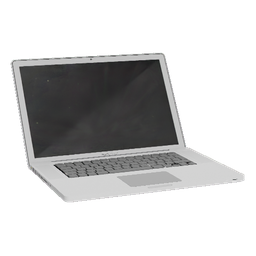}};
  \node[anchor=south east,font=\scriptsize,fill=white,fill opacity=0.8,text opacity=1,inner sep=2pt]
      at (img.south east) {laptop};
\end{tikzpicture} 
 \begin{tikzpicture}
  \node[inner sep=0] (img) {\includegraphics[width=0.115\textwidth]{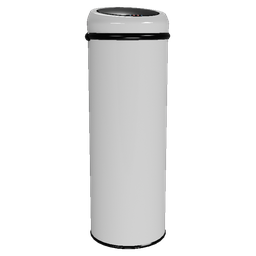}};
  \node[anchor=south east,font=\scriptsize,fill=white,fill opacity=0.8,text opacity=1,inner sep=2pt]
      at (img.south east) {lidded bin};
\end{tikzpicture} 
 \begin{tikzpicture}
  \node[inner sep=0] (img) {\includegraphics[width=0.115\textwidth]{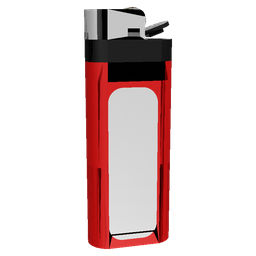}};
  \node[anchor=south east,font=\scriptsize,fill=white,fill opacity=0.8,text opacity=1,inner sep=2pt]
      at (img.south east) {lighter};
\end{tikzpicture} 
 \begin{tikzpicture}
  \node[inner sep=0] (img) {\includegraphics[width=0.115\textwidth]{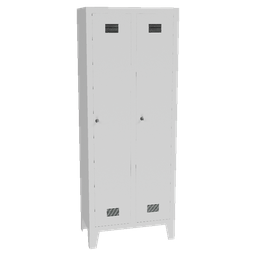}};
  \node[anchor=south east,font=\scriptsize,fill=white,fill opacity=0.8,text opacity=1,inner sep=2pt]
      at (img.south east) {locker};
\end{tikzpicture} 
 \begin{tikzpicture}
  \node[inner sep=0] (img) {\includegraphics[width=0.115\textwidth]{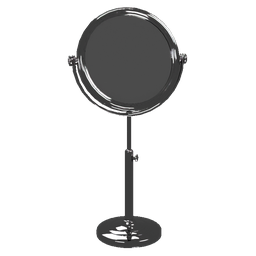}};
  \node[anchor=south east,font=\scriptsize,fill=white,fill opacity=0.8,text opacity=1,inner sep=2pt]
      at (img.south east) {makeup mirror};
\end{tikzpicture} \\
\begin{tikzpicture}
  \node[inner sep=0] (img) {\includegraphics[width=0.115\textwidth]{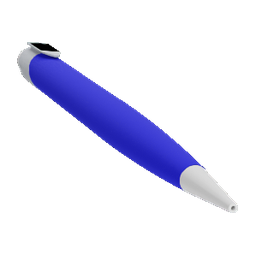}};
  \node[anchor=south east,font=\scriptsize,fill=white,fill opacity=0.8,text opacity=1,inner sep=2pt]
      at (img.south east) {mechanical pen};
\end{tikzpicture} 
 \begin{tikzpicture}
  \node[inner sep=0] (img) {\includegraphics[width=0.115\textwidth]{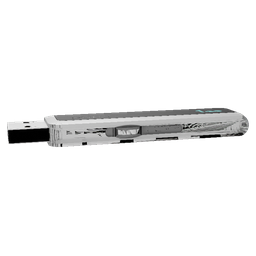}};
  \node[anchor=south east,font=\scriptsize,fill=white,fill opacity=0.8,text opacity=1,inner sep=2pt]
      at (img.south east) {memory stick};
\end{tikzpicture} 
 \begin{tikzpicture}
  \node[inner sep=0] (img) {\includegraphics[width=0.115\textwidth]{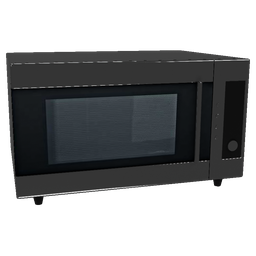}};
  \node[anchor=south east,font=\scriptsize,fill=white,fill opacity=0.8,text opacity=1,inner sep=2pt]
      at (img.south east) {microwave};
\end{tikzpicture} 
 \begin{tikzpicture}
  \node[inner sep=0] (img) {\includegraphics[width=0.115\textwidth]{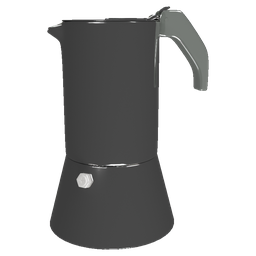}};
  \node[anchor=south east,font=\scriptsize,fill=white,fill opacity=0.8,text opacity=1,inner sep=2pt]
      at (img.south east) {moka pot};
\end{tikzpicture} 
 \begin{tikzpicture}
  \node[inner sep=0] (img) {\includegraphics[width=0.115\textwidth]{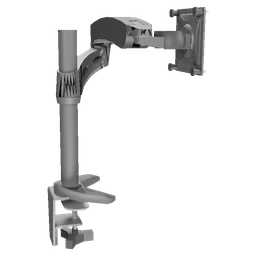}};
  \node[anchor=south east,font=\scriptsize,fill=white,fill opacity=0.8,text opacity=1,inner sep=2pt]
      at (img.south east) {monitor arm};
\end{tikzpicture} 
 \begin{tikzpicture}
  \node[inner sep=0] (img) {\includegraphics[width=0.115\textwidth]{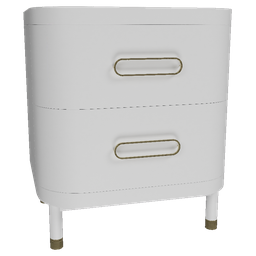}};
  \node[anchor=south east,font=\scriptsize,fill=white,fill opacity=0.8,text opacity=1,inner sep=2pt]
      at (img.south east) {nightstand};
\end{tikzpicture} 
 \begin{tikzpicture}
  \node[inner sep=0] (img) {\includegraphics[width=0.115\textwidth]{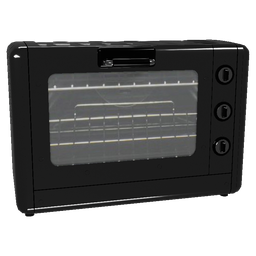}};
  \node[anchor=south east,font=\scriptsize,fill=white,fill opacity=0.8,text opacity=1,inner sep=2pt]
      at (img.south east) {oven};
\end{tikzpicture} 
 \begin{tikzpicture}
  \node[inner sep=0] (img) {\includegraphics[width=0.115\textwidth]{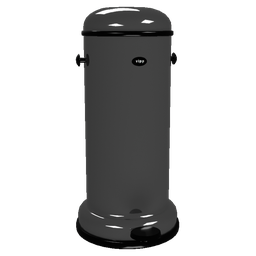}};
  \node[anchor=south east,font=\scriptsize,fill=white,fill opacity=0.8,text opacity=1,inner sep=2pt]
      at (img.south east) {pedal bin};
\end{tikzpicture} \\
\begin{tikzpicture}
  \node[inner sep=0] (img) {\includegraphics[width=0.115\textwidth]{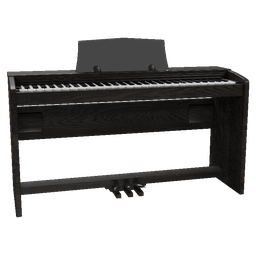}};
  \node[anchor=south east,font=\scriptsize,fill=white,fill opacity=0.8,text opacity=1,inner sep=2pt]
      at (img.south east) {piano};
\end{tikzpicture} 
 \begin{tikzpicture}
  \node[inner sep=0] (img) {\includegraphics[width=0.115\textwidth]{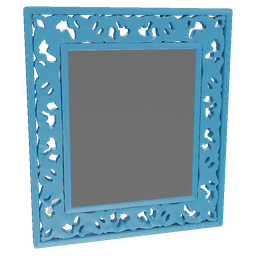}};
  \node[anchor=south east,font=\scriptsize,fill=white,fill opacity=0.8,text opacity=1,inner sep=2pt]
      at (img.south east) {picture frame};
\end{tikzpicture} 
 \begin{tikzpicture}
  \node[inner sep=0] (img) {\includegraphics[width=0.115\textwidth]{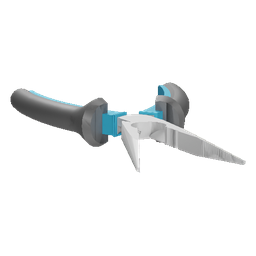}};
  \node[anchor=south east,font=\scriptsize,fill=white,fill opacity=0.8,text opacity=1,inner sep=2pt]
      at (img.south east) {pliers};
\end{tikzpicture} 
 \begin{tikzpicture}
  \node[inner sep=0] (img) {\includegraphics[width=0.115\textwidth]{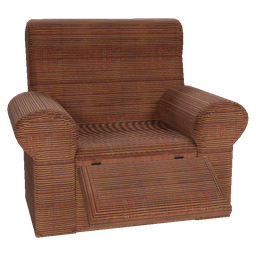}};
  \node[anchor=south east,font=\scriptsize,fill=white,fill opacity=0.8,text opacity=1,inner sep=2pt]
      at (img.south east) {recliner chair};
\end{tikzpicture} 
 \begin{tikzpicture}
  \node[inner sep=0] (img) {\includegraphics[width=0.115\textwidth]{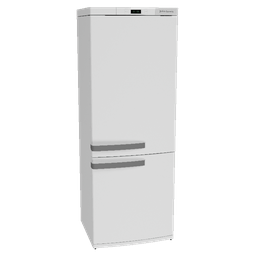}};
  \node[anchor=south east,font=\scriptsize,fill=white,fill opacity=0.8,text opacity=1,inner sep=2pt]
      at (img.south east) {refrigerator};
\end{tikzpicture} 
 \begin{tikzpicture}
  \node[inner sep=0] (img) {\includegraphics[width=0.115\textwidth]{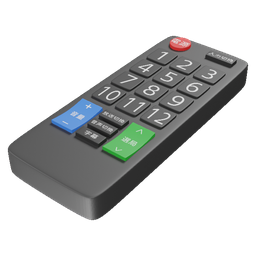}};
  \node[anchor=south east,font=\scriptsize,fill=white,fill opacity=0.8,text opacity=1,inner sep=2pt]
      at (img.south east) {remote};
\end{tikzpicture} 
 \begin{tikzpicture}
  \node[inner sep=0] (img) {\includegraphics[width=0.115\textwidth]{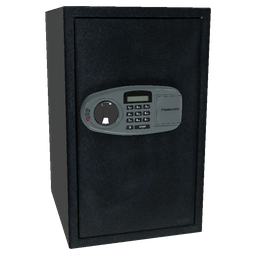}};
  \node[anchor=south east,font=\scriptsize,fill=white,fill opacity=0.8,text opacity=1,inner sep=2pt]
      at (img.south east) {safe};
\end{tikzpicture} 
 \begin{tikzpicture}
  \node[inner sep=0] (img) {\includegraphics[width=0.115\textwidth]{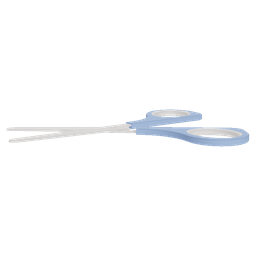}};
  \node[anchor=south east,font=\scriptsize,fill=white,fill opacity=0.8,text opacity=1,inner sep=2pt]
      at (img.south east) {scissors};
\end{tikzpicture} \\
\begin{tikzpicture}
  \node[inner sep=0] (img) {\includegraphics[width=0.115\textwidth]{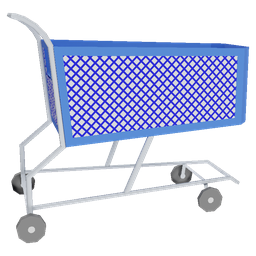}};
  \node[anchor=south east,font=\scriptsize,fill=white,fill opacity=0.8,text opacity=1,inner sep=2pt]
      at (img.south east) {shopping cart};
\end{tikzpicture} 
 \begin{tikzpicture}
  \node[inner sep=0] (img) {\includegraphics[width=0.115\textwidth]{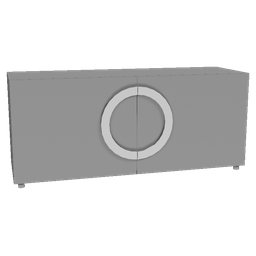}};
  \node[anchor=south east,font=\scriptsize,fill=white,fill opacity=0.8,text opacity=1,inner sep=2pt]
      at (img.south east) {sideboard};
\end{tikzpicture} 
 \begin{tikzpicture}
  \node[inner sep=0] (img) {\includegraphics[width=0.115\textwidth]{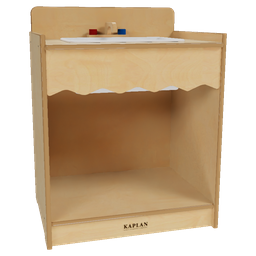}};
  \node[anchor=south east,font=\scriptsize,fill=white,fill opacity=0.8,text opacity=1,inner sep=2pt]
      at (img.south east) {sink cabinet};
\end{tikzpicture} 
 \begin{tikzpicture}
  \node[inner sep=0] (img) {\includegraphics[width=0.115\textwidth]{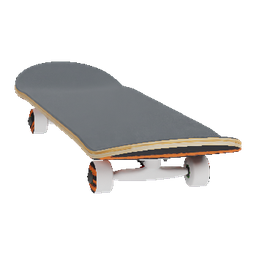}};
  \node[anchor=south east,font=\scriptsize,fill=white,fill opacity=0.8,text opacity=1,inner sep=2pt]
      at (img.south east) {skateboard};
\end{tikzpicture} 
 \begin{tikzpicture}
  \node[inner sep=0] (img) {\includegraphics[width=0.115\textwidth]{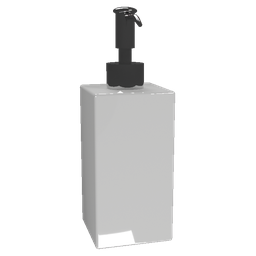}};
  \node[anchor=south east,font=\scriptsize,fill=white,fill opacity=0.8,text opacity=1,inner sep=2pt]
      at (img.south east) {soap dispenser};
\end{tikzpicture} 
 \begin{tikzpicture}
  \node[inner sep=0] (img) {\includegraphics[width=0.115\textwidth]{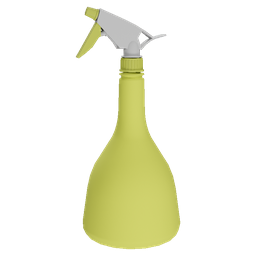}};
  \node[anchor=south east,font=\scriptsize,fill=white,fill opacity=0.8,text opacity=1,inner sep=2pt]
      at (img.south east) {spray bottle};
\end{tikzpicture} 
 \begin{tikzpicture}
  \node[inner sep=0] (img) {\includegraphics[width=0.115\textwidth]{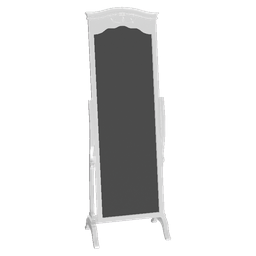}};
  \node[anchor=south east,font=\scriptsize,fill=white,fill opacity=0.8,text opacity=1,inner sep=2pt]
      at (img.south east) {standing mirror};
\end{tikzpicture} 
 \begin{tikzpicture}
  \node[inner sep=0] (img) {\includegraphics[width=0.115\textwidth]{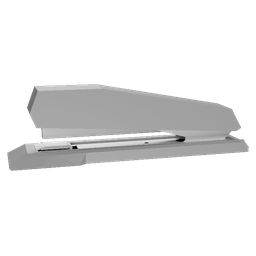}};
  \node[anchor=south east,font=\scriptsize,fill=white,fill opacity=0.8,text opacity=1,inner sep=2pt]
      at (img.south east) {stapler};
\end{tikzpicture} \\
\begin{tikzpicture}
  \node[inner sep=0] (img) {\includegraphics[width=0.115\textwidth]{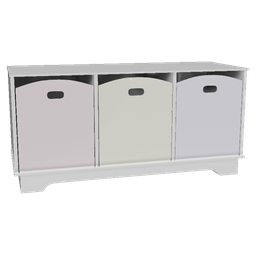}};
  \node[anchor=south east,font=\scriptsize,fill=white,fill opacity=0.8,text opacity=1,inner sep=2pt]
      at (img.south east) {storage bench};
\end{tikzpicture} 
 \begin{tikzpicture}
  \node[inner sep=0] (img) {\includegraphics[width=0.115\textwidth]{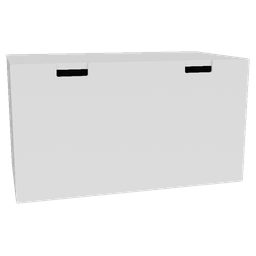}};
  \node[anchor=south east,font=\scriptsize,fill=white,fill opacity=0.8,text opacity=1,inner sep=2pt]
      at (img.south east) {storage box};
\end{tikzpicture} 
 \begin{tikzpicture}
  \node[inner sep=0] (img) {\includegraphics[width=0.115\textwidth]{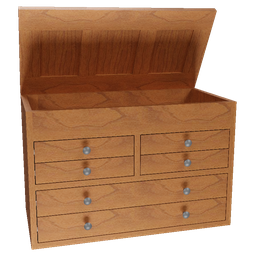}};
  \node[anchor=south east,font=\scriptsize,fill=white,fill opacity=0.8,text opacity=1,inner sep=2pt]
      at (img.south east) {storage chest};
\end{tikzpicture} 
 \begin{tikzpicture}
  \node[inner sep=0] (img) {\includegraphics[width=0.115\textwidth]{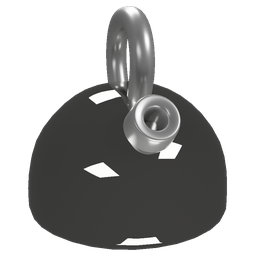}};
  \node[anchor=south east,font=\scriptsize,fill=white,fill opacity=0.8,text opacity=1,inner sep=2pt]
      at (img.south east) {stovetop kettle};
\end{tikzpicture} 
 \begin{tikzpicture}
  \node[inner sep=0] (img) {\includegraphics[width=0.115\textwidth]{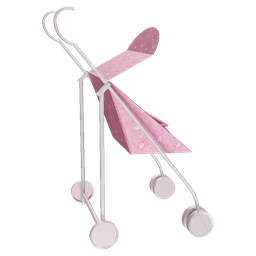}};
  \node[anchor=south east,font=\scriptsize,fill=white,fill opacity=0.8,text opacity=1,inner sep=2pt]
      at (img.south east) {stroller};
\end{tikzpicture} 
 \begin{tikzpicture}
  \node[inner sep=0] (img) {\includegraphics[width=0.115\textwidth]{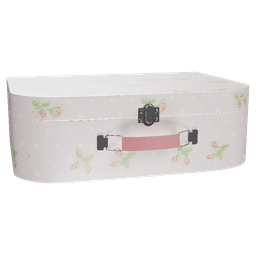}};
  \node[anchor=south east,font=\scriptsize,fill=white,fill opacity=0.8,text opacity=1,inner sep=2pt]
      at (img.south east) {suitcase};
\end{tikzpicture} 
 \begin{tikzpicture}
  \node[inner sep=0] (img) {\includegraphics[width=0.115\textwidth]{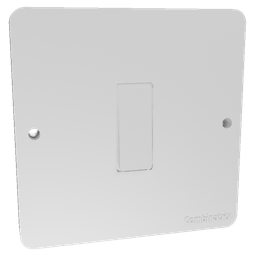}};
  \node[anchor=south east,font=\scriptsize,fill=white,fill opacity=0.8,text opacity=1,inner sep=2pt]
      at (img.south east) {switch};
\end{tikzpicture} 
 \begin{tikzpicture}
  \node[inner sep=0] (img) {\includegraphics[width=0.115\textwidth]{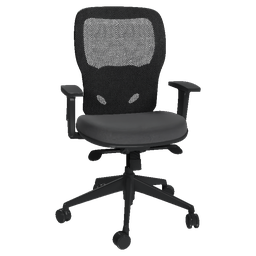}};
  \node[anchor=south east,font=\scriptsize,fill=white,fill opacity=0.8,text opacity=1,inner sep=2pt]
      at (img.south east) {swivel chair};
\end{tikzpicture} \\
\begin{tikzpicture}
  \node[inner sep=0] (img) {\includegraphics[width=0.115\textwidth]{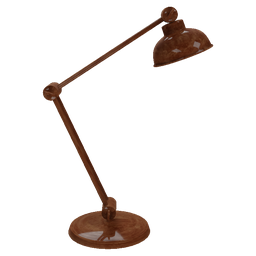}};
  \node[anchor=south east,font=\scriptsize,fill=white,fill opacity=0.8,text opacity=1,inner sep=2pt]
      at (img.south east) {table lamp};
\end{tikzpicture} 
 \begin{tikzpicture}
  \node[inner sep=0] (img) {\includegraphics[width=0.115\textwidth]{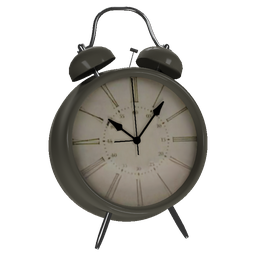}};
  \node[anchor=south east,font=\scriptsize,fill=white,fill opacity=0.8,text opacity=1,inner sep=2pt]
      at (img.south east) {tabletop clock};
\end{tikzpicture} 
 \begin{tikzpicture}
  \node[inner sep=0] (img) {\includegraphics[width=0.115\textwidth]{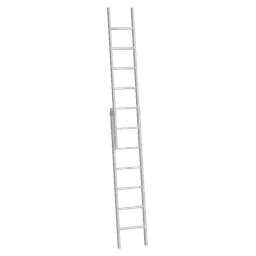}};
  \node[anchor=south east,font=\scriptsize,fill=white,fill opacity=0.8,text opacity=1,inner sep=2pt]
      at (img.south east) {telescoping ladder};
\end{tikzpicture} 
 \begin{tikzpicture}
  \node[inner sep=0] (img) {\includegraphics[width=0.115\textwidth]{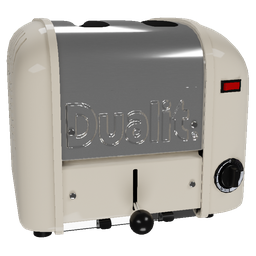}};
  \node[anchor=south east,font=\scriptsize,fill=white,fill opacity=0.8,text opacity=1,inner sep=2pt]
      at (img.south east) {toaster};
\end{tikzpicture} 
 \begin{tikzpicture}
  \node[inner sep=0] (img) {\includegraphics[width=0.115\textwidth]{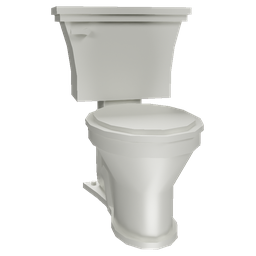}};
  \node[anchor=south east,font=\scriptsize,fill=white,fill opacity=0.8,text opacity=1,inner sep=2pt]
      at (img.south east) {toilet};
\end{tikzpicture} 
 \begin{tikzpicture}
  \node[inner sep=0] (img) {\includegraphics[width=0.115\textwidth]{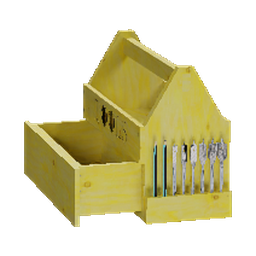}};
  \node[anchor=south east,font=\scriptsize,fill=white,fill opacity=0.8,text opacity=1,inner sep=2pt]
      at (img.south east) {tool box};
\end{tikzpicture} 
 \begin{tikzpicture}
  \node[inner sep=0] (img) {\includegraphics[width=0.115\textwidth]{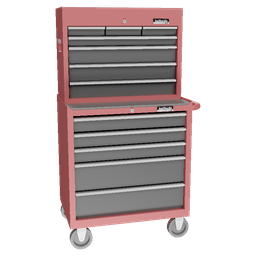}};
  \node[anchor=south east,font=\scriptsize,fill=white,fill opacity=0.8,text opacity=1,inner sep=2pt]
      at (img.south east) {tool cabinet};
\end{tikzpicture} 
 \begin{tikzpicture}
  \node[inner sep=0] (img) {\includegraphics[width=0.115\textwidth]{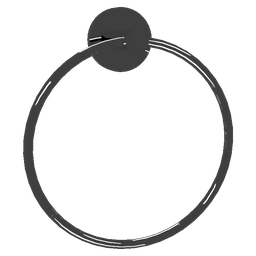}};
  \node[anchor=south east,font=\scriptsize,fill=white,fill opacity=0.8,text opacity=1,inner sep=2pt]
      at (img.south east) {towel ring};
\end{tikzpicture} \\
\begin{tikzpicture}
  \node[inner sep=0] (img) {\includegraphics[width=0.115\textwidth]{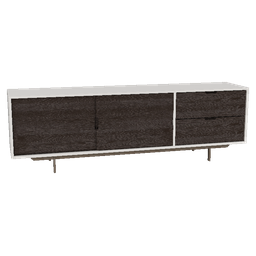}};
  \node[anchor=south east,font=\scriptsize,fill=white,fill opacity=0.8,text opacity=1,inner sep=2pt]
      at (img.south east) {tv stand};
\end{tikzpicture} 
 \begin{tikzpicture}
  \node[inner sep=0] (img) {\includegraphics[width=0.115\textwidth]{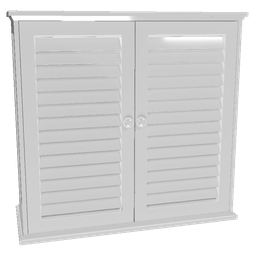}};
  \node[anchor=south east,font=\scriptsize,fill=white,fill opacity=0.8,text opacity=1,inner sep=2pt]
      at (img.south east) {wall cabinet};
\end{tikzpicture} 
 \begin{tikzpicture}
  \node[inner sep=0] (img) {\includegraphics[width=0.115\textwidth]{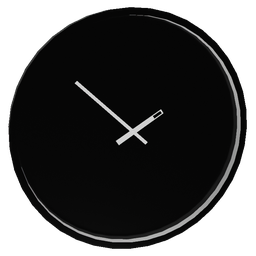}};
  \node[anchor=south east,font=\scriptsize,fill=white,fill opacity=0.8,text opacity=1,inner sep=2pt]
      at (img.south east) {wall clock};
\end{tikzpicture} 
 \begin{tikzpicture}
  \node[inner sep=0] (img) {\includegraphics[width=0.115\textwidth]{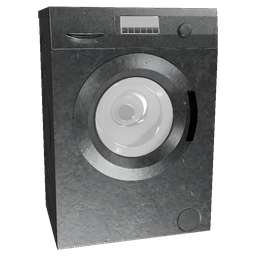}};
  \node[anchor=south east,font=\scriptsize,fill=white,fill opacity=0.8,text opacity=1,inner sep=2pt]
      at (img.south east) {washing machine};
\end{tikzpicture} 
 \begin{tikzpicture}
  \node[inner sep=0] (img) {\includegraphics[width=0.115\textwidth]{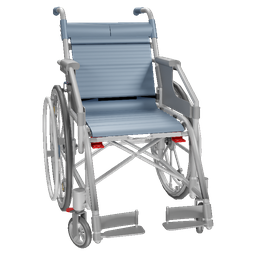}};
  \node[anchor=south east,font=\scriptsize,fill=white,fill opacity=0.8,text opacity=1,inner sep=2pt]
      at (img.south east) {wheelchair};
\end{tikzpicture} 
 \begin{tikzpicture}
  \node[inner sep=0] (img) {\includegraphics[width=0.115\textwidth]{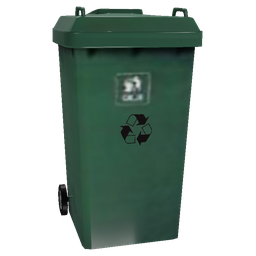}};
  \node[anchor=south east,font=\scriptsize,fill=white,fill opacity=0.8,text opacity=1,inner sep=2pt]
      at (img.south east) {wheelie bin};
\end{tikzpicture} 
 \begin{tikzpicture}
  \node[inner sep=0] (img) {\includegraphics[width=0.115\textwidth]{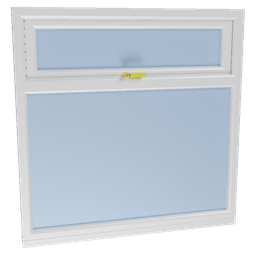}};
  \node[anchor=south east,font=\scriptsize,fill=white,fill opacity=0.8,text opacity=1,inner sep=2pt]
      at (img.south east) {window};
\end{tikzpicture} 
 \begin{tikzpicture}
  \node[inner sep=0] (img) {\includegraphics[width=0.115\textwidth]{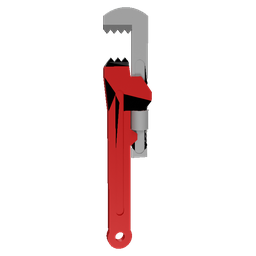}};
  \node[anchor=south east,font=\scriptsize,fill=white,fill opacity=0.8,text opacity=1,inner sep=2pt]
      at (img.south east) {wrench};
\end{tikzpicture} \\
}
\caption{Preview of all Artiverse categories.} 
\end{figure}
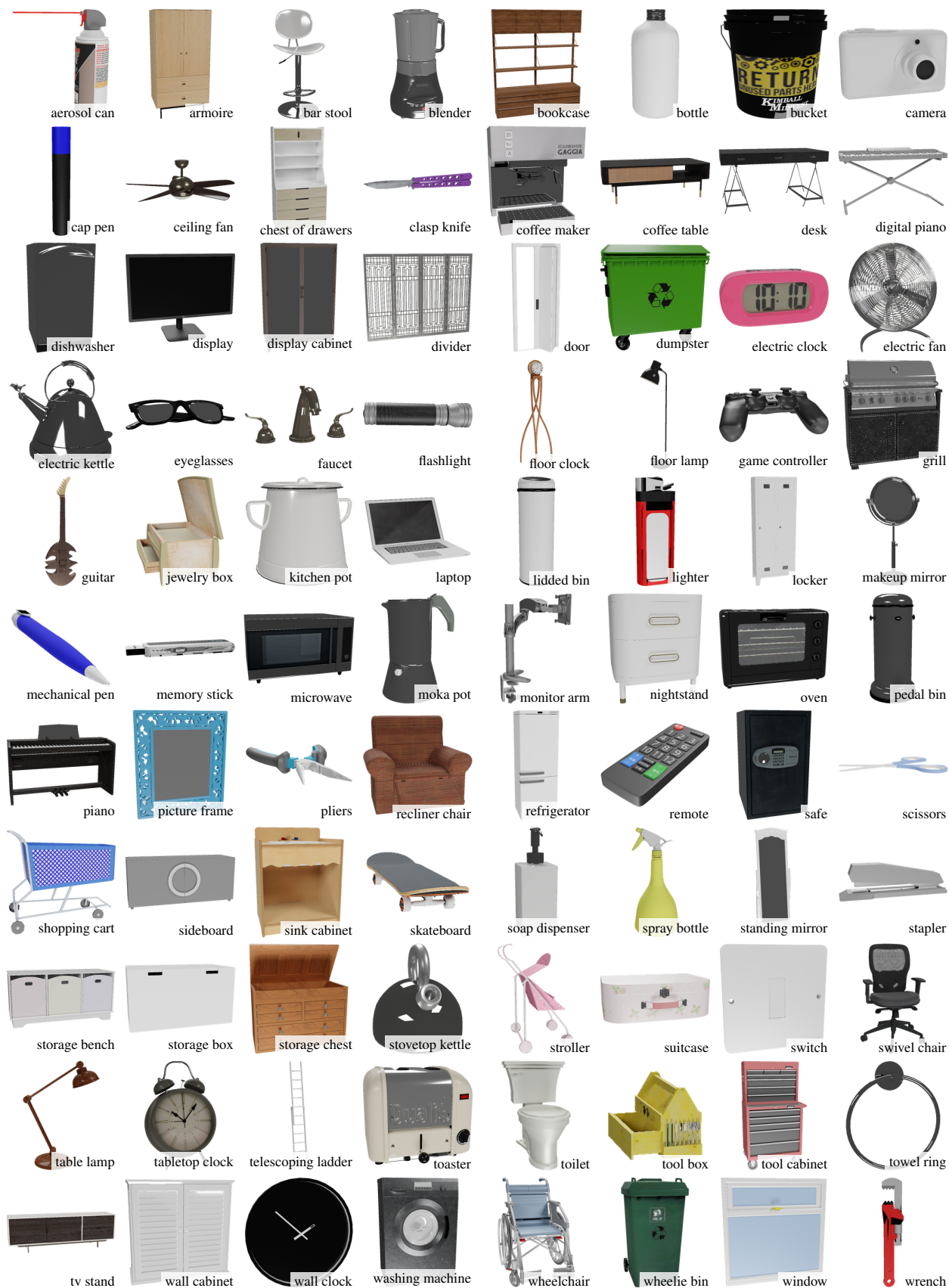

\twocolumn

%% file: main.bbl
\begin{thebibliography}{59}
\providecommand{\natexlab}[1]{#1}
\providecommand{\url}[1]{\texttt{#1}}
\expandafter\ifx\csname urlstyle\endcsname\relax
  \providecommand{\doi}[1]{doi: #1}\else
  \providecommand{\doi}{doi: \begingroup \urlstyle{rm}\Url}\fi

\bibitem[Abdelreheem et~al.(2023)Abdelreheem, Skorokhodov, Ovsjanikov, and Wonka]{abdelreheem2023satr}
Ahmed Abdelreheem, Ivan Skorokhodov, Maks Ovsjanikov, and Peter Wonka.
\newblock {SATR}: Zero-shot semantic segmentation of {3D} shapes.
\newblock In \emph{ICCV}, pages 15166--15179, 2023.

\bibitem[Ahmed et~al.(2025)Ahmed, Li, Prajapati, and Elhoseiny]{ahmed20253dcompat200}
Mahmoud Ahmed, Xiang Li, Arpit Prajapati, and Mohamed Elhoseiny.
\newblock {3DCoMPaT200}: Language-grounded compositional understanding of parts and materials of 3d shapes.
\newblock \emph{arXiv preprint arXiv:2501.06785}, 2025.

\bibitem[Authors(2024)]{Genesis}
Genesis Authors.
\newblock Genesis: A generative and universal physics engine for robotics and beyond, 2024.

\bibitem[Cao et~al.(2025)Cao, Chen, Pan, and Liu]{cao2025physx}
Ziang Cao, Zhaoxi Chen, Linag Pan, and Ziwei Liu.
\newblock {PhysX-3D}: Physical-grounded {3D} asset generation.
\newblock \emph{arXiv preprint arXiv:2507.12465}, 2025.

\bibitem[Chang et~al.(2015)Chang, Funkhouser, Guibas, Hanrahan, Huang, Li, Savarese, Savva, Song, Su, et~al.]{chang2015shapenet}
Angel~X Chang, Thomas Funkhouser, Leonidas Guibas, Pat Hanrahan, Qixing Huang, Zimo Li, Silvio Savarese, Manolis Savva, Shuran Song, Hao Su, et~al.
\newblock Shapenet: An information-rich {3D} model repository.
\newblock \emph{arXiv preprint arXiv:1512.03012}, 2015.

\bibitem[Collins et~al.(2022)Collins, Goel, Deng, Luthra, Xu, Gundogdu, Zhang, Vicente, Dideriksen, Arora, et~al.]{collins2022abo}
Jasmine Collins, Shubham Goel, Kenan Deng, Achleshwar Luthra, Leon Xu, Erhan Gundogdu, Xi Zhang, Tomas F~Yago Vicente, Thomas Dideriksen, Himanshu Arora, et~al.
\newblock {ABO}: Dataset and benchmarks for real-world {3D} object understanding.
\newblock In \emph{CVPR}, pages 21126--21136, 2022.

\bibitem[Decatur et~al.(2023)Decatur, Lang, and Hanocka]{decatur20233d}
Dale Decatur, Itai Lang, and Rana Hanocka.
\newblock {3D} highlighter: Localizing regions on {3D} shapes via text descriptions.
\newblock In \emph{CVPR}, pages 20930--20939, 2023.

\bibitem[Decatur et~al.(2024)Decatur, Lang, Aberman, and Hanocka]{decatur20243d}
Dale Decatur, Itai Lang, Kfir Aberman, and Rana Hanocka.
\newblock {3D} paintbrush: Local stylization of {3D} shapes with cascaded score distillation.
\newblock In \emph{CVPR}, pages 4473--4483, 2024.

\bibitem[Deitke et~al.(2023)Deitke, Schwenk, Salvador, Weihs, Michel, VanderBilt, Schmidt, Ehsani, Kembhavi, and Farhadi]{deitke2023objaverse}
Matt Deitke, Dustin Schwenk, Jordi Salvador, Luca Weihs, Oscar Michel, Eli VanderBilt, Ludwig Schmidt, Kiana Ehsani, Aniruddha Kembhavi, and Ali Farhadi.
\newblock Objaverse: A universe of annotated {3D} objects.
\newblock In \emph{CVPR}, pages 13142--13153, 2023.

\bibitem[Deitke et~al.(2024)Deitke, Liu, Wallingford, Ngo, Michel, Kusupati, Fan, Laforte, Voleti, Gadre, et~al.]{deitke2024objaversexl}
Matt Deitke, Ruoshi Liu, Matthew Wallingford, Huong Ngo, Oscar Michel, Aditya Kusupati, Alan Fan, Christian Laforte, Vikram Voleti, Samir~Yitzhak Gadre, et~al.
\newblock {Objaverse-XL}: A universe of {10M+ 3D} objects.
\newblock \emph{NeurIPS}, 36, 2024.

\bibitem[Deng et~al.(2025)Deng, Subr, and Bilen]{deng2025paoli}
Jianning Deng, Kartic Subr, and Hakan Bilen.
\newblock {PAOLI}: Pose-free articulated object learning from sparse-view images.
\newblock \emph{arXiv preprint arXiv:2509.04276}, 2025.

\bibitem[Fisher et~al.(2012)Fisher, Ritchie, Savva, Funkhouser, and Hanrahan]{fisher2012example}
Matthew Fisher, Daniel Ritchie, Manolis Savva, Thomas Funkhouser, and Pat Hanrahan.
\newblock Example-based synthesis of {3D} object arrangements.
\newblock \emph{ACM TOG}, 31\penalty0 (6):\penalty0 1--11, 2012.

\bibitem[Fu et~al.(2021)Fu, Jia, Gao, Gong, Zhao, Maybank, and Tao]{fu20213d}
Huan Fu, Rongfei Jia, Lin Gao, Mingming Gong, Binqiang Zhao, Steve Maybank, and Dacheng Tao.
\newblock {3D-Future: 3D} furniture shape with texture.
\newblock \emph{International Journal of Computer Vision}, 129:\penalty0 3313--3337, 2021.

\bibitem[Gao et~al.(2025)Gao, Siddiqui, Li, and Dai]{gao2025meshart}
Daoyi Gao, Yawar Siddiqui, Lei Li, and Angela Dai.
\newblock {MeshArt}: Generating articulated meshes with structure-guided transformers.
\newblock In \emph{CVPR}, pages 618--627, 2025.

\bibitem[Geng et~al.(2023)Geng, Xu, Zhao, Xu, Yi, Huang, and Wang]{geng2023gapartnet}
Haoran Geng, Helin Xu, Chengyang Zhao, Chao Xu, Li Yi, Siyuan Huang, and He Wang.
\newblock {GAPartNet}: Cross-category domain-generalizable object perception and manipulation via generalizable and actionable parts.
\newblock In \emph{CVPR}, pages 7081--7091, 2023.

\bibitem[Goyal et~al.(2025)Goyal, Petrov, Andrews, Ben-Shabat, Liu, and Kalogerakis]{goyal2025geopard}
Pradyumn Goyal, Dmitry Petrov, Sheldon Andrews, Yizhak Ben-Shabat, Hsueh-Ti~Derek Liu, and Evangelos Kalogerakis.
\newblock {GEOPARD}: Geometric pretraining for articulation prediction in {3D} shapes.
\newblock \emph{arXiv preprint arXiv:2504.02747}, 2025.

\bibitem[Guo et~al.(2025{\natexlab{a}})Guo, Xin, Liu, Xu, Liu, and Hu]{guo2025articulatedgs}
Junfu Guo, Yu Xin, Gaoyi Liu, Kai Xu, Ligang Liu, and Ruizhen Hu.
\newblock {ArticulatedGS}: Self-supervised digital twin modeling of articulated objects using 3d gaussian splatting.
\newblock In \emph{CVPR}, pages 27144--27153, 2025{\natexlab{a}}.

\bibitem[Guo et~al.(2025{\natexlab{b}})Guo, Zordan, Andrews, Matusik, Agrawala, and Liu]{guo2025kinematic}
Minghao Guo, Victor Zordan, Sheldon Andrews, Wojciech Matusik, Maneesh Agrawala, and Hsueh-Ti~Derek Liu.
\newblock Kinematic kitbashing for modeling functional articulated objects.
\newblock \emph{arXiv preprint arXiv:2510.13048}, 2025{\natexlab{b}}.

\bibitem[Iliash et~al.(2024)Iliash, Jiang, Zhang, Savva, and Chang]{iliash2024s2o}
Denys Iliash, Hanxiao Jiang, Yiming Zhang, Manolis Savva, and Angel~X Chang.
\newblock {S2O}: Static to openable enhancement for articulated {3D} objects.
\newblock \emph{arXiv preprint arXiv:2409.18896}, 2024.

\bibitem[Jiang et~al.(2022)Jiang, Mao, Savva, and Chang]{jiang2022opd}
Hanxiao Jiang, Yongsen Mao, Manolis Savva, and Angel~X Chang.
\newblock {OPD}: Single-view 3d openable part detection.
\newblock In \emph{ECCV}, pages 410--426, 2022.

\bibitem[Jin et~al.(2025)Jin, Che, Zhao, Wu, Zhang, Zhao, Liu, Zhang, Ju, Tian, et~al.]{jin2025artvip}
Zhao Jin, Zhengping Che, Zhen Zhao, Kun Wu, Yuheng Zhang, Yinuo Zhao, Zehui Liu, Qiang Zhang, Xiaozhu Ju, Jing Tian, et~al.
\newblock {ArtVIP}: Articulated digital assets of visual realism, modular interaction, and physical fidelity for robot learning.
\newblock \emph{arXiv preprint arXiv:2506.04941}, 2025.

\bibitem[Joshi et~al.(2025)Joshi, Han, Nugent, Zuo, Liu, Wen, Alexandropoulos, Sun, Raistrick, Liu, et~al.]{joshi2025infinigen}
Abhishek Joshi, Beining Han, Jack Nugent, Yiming Zuo, Jonathan Liu, Hongyu Wen, Stamatis Alexandropoulos, Tao Sun, Alexander Raistrick, Gaowen Liu, et~al.
\newblock {Infinigen-Sim}: Procedural generation of articulated simulation assets.
\newblock \emph{arXiv preprint arXiv:2505.10755}, 2025.

\bibitem[Kerr et~al.(2024)Kerr, Kim, Wu, Yi, Wang, Goldberg, and Kanazawa]{kerr2024robot}
Justin Kerr, Chung~Min Kim, Mingxuan Wu, Brent Yi, Qianqian Wang, Ken Goldberg, and Angjoo Kanazawa.
\newblock Robot see robot do: Imitating articulated object manipulation with monocular 4d reconstruction.
\newblock \emph{arXiv preprint arXiv:2409.18121}, 2024.

\bibitem[Khanna et~al.(2024)Khanna, Mao, Jiang, Haresh, Shacklett, Batra, Clegg, Undersander, Chang, and Savva]{khanna2024habitat}
Mukul Khanna, Yongsen Mao, Hanxiao Jiang, Sanjay Haresh, Brennan Shacklett, Dhruv Batra, Alexander Clegg, Eric Undersander, Angel~X Chang, and Manolis Savva.
\newblock Habitat synthetic scenes dataset ({HSSD}-200): An analysis of {3D} scene scale and realism tradeoffs for objectgoal navigation.
\newblock In \emph{CVPR}, pages 16384--16393, 2024.

\bibitem[Kim and Sung(2024)]{kim2024partstad}
Hyunjin Kim and Minhyuk Sung.
\newblock {PartSTAD: 2D-to-3D} part segmentation task adaptation.
\newblock In \emph{ECCV}, pages 422--439, 2024.

\bibitem[Kreber and Stueckler(2025)]{kreber2025guiding}
Jens~U Kreber and Joerg Stueckler.
\newblock Guiding diffusion-based articulated object generation by partial point cloud alignment and physical plausibility constraints.
\newblock In \emph{ICCV}, pages 3206--3214, 2025.

\bibitem[Le et~al.(2024)Le, Xie, Liang, Wang, Yang, Ma, Vedder, Krishna, Jayaraman, and Eaton]{le2024articulate}
Long Le, Jason Xie, William Liang, Hung-Ju Wang, Yue Yang, Yecheng~Jason Ma, Kyle Vedder, Arjun Krishna, Dinesh Jayaraman, and Eric Eaton.
\newblock {Articulate-Anything}: Automatic modeling of articulated objects via a vision-language foundation model.
\newblock \emph{arXiv preprint arXiv:2410.13882}, 2024.

\bibitem[Lee et~al.(2025)Lee, Zhang, and Chang]{lee2025duoduo}
Han-Hung Lee, Yiming Zhang, and Angel~X Chang.
\newblock Duoduo {CLIP}: Efficient {3D} understanding with multi-view images.
\newblock In \emph{ICLR}, 2025.

\bibitem[Lei et~al.(2023)Lei, Deng, Shen, Guibas, and Daniilidis]{lei2023nap}
Jiahui Lei, Congyue Deng, Bokui Shen, Leonidas Guibas, and Kostas Daniilidis.
\newblock Nap: Neural 3d articulation prior.
\newblock \emph{arXiv preprint arXiv:2305.16315}, 2023.

\bibitem[Lian et~al.(2025)Lian, Yu, Liang, Wang, Luo, Chen, Zhou, Tang, Xu, Lyu, et~al.]{lian2025infinite}
Xinyu Lian, Zichao Yu, Ruiming Liang, Yitong Wang, Li~Ray Luo, Kaixu Chen, Yuanzhen Zhou, Qihong Tang, Xudong Xu, Zhaoyang Lyu, et~al.
\newblock Infinite mobility: Scalable high-fidelity synthesis of articulated objects via procedural generation.
\newblock \emph{arXiv preprint arXiv:2503.13424}, 2025.

\bibitem[Liu et~al.(2023{\natexlab{a}})Liu, Mahdavi-Amiri, and Savva]{liu2023paris}
Jiayi Liu, Ali Mahdavi-Amiri, and Manolis Savva.
\newblock {PARIS}: Part-level reconstruction and motion analysis for articulated objects.
\newblock In \emph{ICCV}, pages 352--363, 2023{\natexlab{a}}.

\bibitem[Liu et~al.(2024{\natexlab{a}})Liu, Iliash, Chang, Savva, and Mahdavi-Amiri]{liu2024singapo}
Jiayi Liu, Denys Iliash, Angel~X Chang, Manolis Savva, and Ali Mahdavi-Amiri.
\newblock {SINGAPO}: Single image controlled generation of articulated parts in objects.
\newblock \emph{arXiv preprint arXiv:2410.16499}, 2024{\natexlab{a}}.

\bibitem[Liu et~al.(2024{\natexlab{b}})Liu, Tam, Mahdavi-Amiri, and Savva]{liu2024cage}
Jiayi Liu, Hou In~Ivan Tam, Ali Mahdavi-Amiri, and Manolis Savva.
\newblock {CAGE}: Controllable articulation generation.
\newblock In \emph{CVPR}, pages 17880--17889, 2024{\natexlab{b}}.

\bibitem[Liu et~al.(2022{\natexlab{a}})Liu, Xu, Fu, Qian, Yu, Han, and Lu]{liu2022akb}
Liu Liu, Wenqiang Xu, Haoyuan Fu, Sucheng Qian, Qiaojun Yu, Yang Han, and Cewu Lu.
\newblock {AKB-48}: A real-world articulated object knowledge base.
\newblock In \emph{CVPR}, pages 14809--14818, 2022{\natexlab{a}}.

\bibitem[Liu et~al.(2022{\natexlab{b}})Liu, Xue, Xu, Fu, and Lu]{liu2022toward}
Liu Liu, Han Xue, Wenqiang Xu, Haoyuan Fu, and Cewu Lu.
\newblock Toward real-world category-level articulation pose estimation.
\newblock \emph{IEEE Transactions on Image Processing}, 31:\penalty0 1072--1083, 2022{\natexlab{b}}.

\bibitem[Liu et~al.(2023{\natexlab{b}})Liu, Zhu, Cai, Han, Ling, Porikli, and Su]{liu2023partslip}
Minghua Liu, Yinhao Zhu, Hong Cai, Shizhong Han, Zhan Ling, Fatih Porikli, and Hao Su.
\newblock {PartSLIP}: Low-shot part segmentation for {3D} point clouds via pretrained image-language models.
\newblock In \emph{CVPR}, pages 21736--21746, 2023{\natexlab{b}}.

\bibitem[Liu et~al.(2025)Liu, Jia, Lu, Ni, Zhu, and Huang]{liu2025building}
Yu Liu, Baoxiong Jia, Ruijie Lu, Junfeng Ni, Song-Chun Zhu, and Siyuan Huang.
\newblock Building interactable replicas of complex articulated objects via gaussian splatting.
\newblock \emph{ArXiv}, abs/2502.19459, 2025.

\bibitem[Luo et~al.(2025)Luo, Geng, Deng, Li, Wang, Jia, Guibas, and Huang]{luo2025physpart}
Rundong Luo, Haoran Geng, Congyue Deng, Puhao Li, Zan Wang, Baoxiong Jia, Leonidas Guibas, and Siyuan Huang.
\newblock {PhysPart}: Physically plausible part completion for interactable objects.
\newblock In \emph{IEEE International Conference on Robotics and Automation (ICRA)}, pages 12386--12393, 2025.

\bibitem[Mart{\'\i}n-Mart{\'\i}n et~al.(2019)Mart{\'\i}n-Mart{\'\i}n, Eppner, and Brock]{martin2019rbo}
Roberto Mart{\'\i}n-Mart{\'\i}n, Clemens Eppner, and Oliver Brock.
\newblock The {RBO} dataset of articulated objects and interactions.
\newblock \emph{The International Journal of Robotics Research}, 38\penalty0 (9):\penalty0 1013--1019, 2019.

\bibitem[Mo et~al.(2019)Mo, Zhu, Chang, Yi, Tripathi, Guibas, and Su]{mo2019partnet}
Kaichun Mo, Shilin Zhu, Angel~X Chang, Li Yi, Subarna Tripathi, Leonidas~J Guibas, and Hao Su.
\newblock {PartNet}: A large-scale benchmark for fine-grained and hierarchical part-level {3D} object understanding.
\newblock In \emph{Proceedings of the IEEE/CVF conference on computer vision and pattern recognition}, pages 909--918, 2019.

\bibitem[Perla et~al.(2025)Perla, Vora, Nag, Mahdavi-Amiri, and Zhang]{perla2025asia}
Sai Raj~Kishore Perla, Aditya Vora, Sauradip Nag, Ali Mahdavi-Amiri, and Hao Zhang.
\newblock {ASIA}: Adaptive {3D} segmentation using few image annotations.
\newblock \emph{ArXiv}, abs/2509.24288, 2025.

\bibitem[Ren et~al.(2024)Ren, Li, Luo, Song, Chen, Liufu, Yang, Zheng, Xu, Huang, et~al.]{ren2024infiniteworld}
Pengzhen Ren, Min Li, Zhen Luo, Xinshuai Song, Ziwei Chen, Weijia Liufu, Yixuan Yang, Hao Zheng, Rongtao Xu, Zitong Huang, et~al.
\newblock {InfiniteWorld}: A unified scalable simulation framework for general visual-language robot interaction.
\newblock \emph{arXiv preprint arXiv:2412.05789}, 2024.

\bibitem[Sun et~al.(2025)Sun, Li, Wei, Xu, Wang, Zhang, and Lu]{sun2025arti}
Jianhua Sun, Yuxuan Li, Jiude Wei, Longfei Xu, Nange Wang, Yining Zhang, and Cewu Lu.
\newblock {Arti-PG}: A toolbox for procedurally synthesizing large-scale and diverse articulated objects with rich annotations.
\newblock In \emph{ICCV}, pages 6396--6405, 2025.

\bibitem[Sun et~al.(2024)Sun, Jiang, Savva, and Chang]{sun2024opdmulti}
Xiaohao Sun, Hanxiao Jiang, Manolis Savva, and Angel Chang.
\newblock {OPDMulti}: Openable part detection for multiple objects.
\newblock In \emph{2024 International Conference on 3D Vision (3DV)}, pages 169--178, 2024.

\bibitem[Thai et~al.(2024)Thai, Wang, Tang, Stojanov, Rehg, and Feiszli]{thai20243}
Anh Thai, Weiyao Wang, Hao Tang, Stefan Stojanov, James~M Rehg, and Matt Feiszli.
\newblock 3$\times$ 2: {3D} object part segmentation by {2D} semantic correspondences.
\newblock In \emph{European Conference on Computer Vision}, pages 149--166. Springer, 2024.

\bibitem[Wang et~al.(2025{\natexlab{a}})Wang, He, Lv, Zhou, Xu, Yu, and Gu]{wang2025partnext}
Penghao Wang, Yiyang He, Xin Lv, Yukai Zhou, Lan Xu, Jingyi Yu, and Jiayuan Gu.
\newblock {PartNeXt}: A next-generation dataset for fine-grained and hierarchical {3D} part understanding.
\newblock \emph{arXiv preprint arXiv:2510.20155}, 2025{\natexlab{a}}.

\bibitem[Wang et~al.(2019)Wang, Zhou, Shi, Chen, Zhao, and Xu]{wang2019shape2motion}
Xiaogang Wang, Bin Zhou, Yahao Shi, Xiaowu Chen, Qinping Zhao, and Kai Xu.
\newblock Shape2motion: Joint analysis of motion parts and attributes from 3d shapes.
\newblock In \emph{CVPR}, pages 8876--8884, 2019.

\bibitem[Wang et~al.(2025{\natexlab{b}})Wang, Liu, Cao, Wu, Qin, Wang, Sui, and Su]{wang2025embodiedgen}
Xinjie Wang, Liu Liu, Yu Cao, Ruiqi Wu, Wenkang Qin, Dehui Wang, Wei Sui, and Zhizhong Su.
\newblock {EmbodiedGen}: Towards a generative 3d world engine for embodied intelligence.
\newblock \emph{ArXiv}, abs/2506.10600, 2025{\natexlab{b}}.

\bibitem[Weng et~al.(2024)Weng, Wen, Tremblay, Blukis, Fox, Guibas, and Birchfield]{weng2024neural}
Yijia Weng, Bowen Wen, Jonathan Tremblay, Valts Blukis, Dieter Fox, Leonidas Guibas, and Stan Birchfield.
\newblock Neural implicit representation for building digital twins of unknown articulated objects.
\newblock In \emph{CVPR}, pages 3141--3150, 2024.

\bibitem[Wu et~al.(2025{\natexlab{a}})Wu, Liu, Zhou, Huang, Song, Yu, Wu, and Lu]{wu2025reartgs}
Di Wu, Liu Liu, Linli Zhou, Anran Huang, Liangtu Song, Qiaojun Yu, Qi Wu, and Cewu Lu.
\newblock {REArtGS}: Reconstructing and generating articulated objects via {3D} gaussian splatting with geometric and motion constraints.
\newblock \emph{ArXiv}, abs/2503.06677, 2025{\natexlab{a}}.

\bibitem[Wu et~al.(2025{\natexlab{b}})Wu, Wang, Liu, Guo, Qiu, Li, Huang, Su, and Cheng]{wu2025dipo}
Ruiqi Wu, Xinjie Wang, Liu Liu, Chunle Guo, Jiaxiong Qiu, Chongyi Li, Lichao Huang, Zhizhong Su, and Ming-Ming Cheng.
\newblock {DIPO}: Dual-state images controlled articulated object generation powered by diverse data.
\newblock \emph{arXiv preprint arXiv:2505.20460}, 2025{\natexlab{b}}.

\bibitem[Xia et~al.(2025)Xia, Su, Memmel, Jain, Yu, Mbiziwo-Tiapo, Farhadi, Gupta, Wang, and Ma]{xia2025drawer}
Hongchi Xia, Entong Su, Marius Memmel, Arhan Jain, Raymond Yu, Numfor Mbiziwo-Tiapo, Ali Farhadi, Abhishek Gupta, Shenlong Wang, and Wei-Chiu Ma.
\newblock {DRAWER}: Digital reconstruction and articulation with environment realism.
\newblock In \emph{Proceedings of the Computer Vision and Pattern Recognition Conference}, pages 21771--21782, 2025.

\bibitem[Xiang et~al.(2020)Xiang, Qin, Mo, Xia, Zhu, Liu, Liu, Jiang, Yuan, Wang, et~al.]{xiang2020sapien}
Fanbo Xiang, Yuzhe Qin, Kaichun Mo, Yikuan Xia, Hao Zhu, Fangchen Liu, Minghua Liu, Hanxiao Jiang, Yifu Yuan, He Wang, et~al.
\newblock {SAPIEN}: A simulated part-based interactive environment.
\newblock In \emph{Proceedings of the IEEE/CVF conference on computer vision and pattern recognition}, pages 11097--11107, 2020.

\bibitem[Xiang et~al.(2025)Xiang, Lv, Xu, Deng, Wang, Zhang, Chen, Tong, and Yang]{xiang2025structured}
Jianfeng Xiang, Zelong Lv, Sicheng Xu, Yu Deng, Ruicheng Wang, Bowen Zhang, Dong Chen, Xin Tong, and Jiaolong Yang.
\newblock Structured 3d latents for scalable and versatile 3d generation.
\newblock In \emph{CVPR}, 2025.

\bibitem[Yan et~al.(2025)Yan, Xu, Li, Ma, Yang, Wang, Zhao, Lai, Zhao, Chen, and Guo]{yan2025xpart}
Xinhao Yan, Jiachen Xu, Yang Li, Changfeng Ma, Yunhan Yang, Chunshi Wang, Zibo Zhao, Zeqiang Lai, Yunfei Zhao, Zhuo Chen, and Chunchao Guo.
\newblock X-part: high fidelity and structure coherent shape decomposition.
\newblock \emph{ArXiv}, abs/2509.08643, 2025.

\bibitem[Yan et~al.(2019)Yan, Hu, Yan, Chen, van Kaick, Zhang, and Huang]{yan2019rpmnet}
Zihao Yan, Ruizhen Hu, Xingguang Yan, Luanmin Chen, Oliver van Kaick, Hao Zhang, and Hui Huang.
\newblock {RPM-Net}: Recurrent prediction of motion and parts from point cloud.
\newblock \emph{ACM Transactions on Graphics (Proceedings of SIGGRAPH Asia)}, 38\penalty0 (6):\penalty0 240:1--240:15, 2019.

\bibitem[Yang et~al.(2024)Yang, Jia, Zhi, and Huang]{yang2024physcene}
Yandan Yang, Baoxiong Jia, Peiyuan Zhi, and Siyuan Huang.
\newblock {PhyScene}: Physically interactable 3d scene synthesis for embodied ai.
\newblock In \emph{CVPR}, pages 16262--16272, 2024.

\bibitem[Zhang et~al.(2025)Zhang, Zhang, Ma, and Cao]{zhang2025texverse}
Yibo Zhang, Li Zhang, Rui Ma, and Nan Cao.
\newblock {TexVerse}: A universe of {3D} objects with high-resolution textures.
\newblock \emph{arXiv preprint arXiv:2508.10868}, 2025.

\bibitem[Zhong et~al.(2024)Zhong, Xu, Li, Xu, Li, Yu, and Gao]{zhong2024meshsegmenter}
Ziming Zhong, Yanyu Xu, Jing Li, Jiale Xu, Zhengxin Li, Chaohui Yu, and Shenghua Gao.
\newblock {Meshsegmenter}: Zero-shot mesh semantic segmentation via texture synthesis.
\newblock In \emph{ECCV}, pages 182--199, 2024.

\end{thebibliography}
